\documentclass{article}

\usepackage{PRIMEarxiv}

\usepackage[utf8]{inputenc} 
\usepackage[T1]{fontenc}    
\usepackage{hyperref}       
\usepackage{url}            
\usepackage{booktabs}       
\usepackage{amsfonts}       
\usepackage{nicefrac}       
\usepackage{microtype}      
\usepackage{lipsum}
\usepackage{fancyhdr}       
\usepackage{graphicx}       
\graphicspath{{media/}}     
\usepackage{xcolor}

\usepackage{multirow}
\usepackage{xpatch}
\usepackage{url}            
\usepackage{booktabs}       
\usepackage{amsfonts}       
\usepackage{nicefrac}       
\usepackage{microtype}      
\usepackage{xcolor}         
\usepackage{wrapfig}
\usepackage{amsmath}
\usepackage{amssymb}
\usepackage{mathtools}
\usepackage{amsthm}
\usepackage{xspace} 
\usepackage{colortbl}
\usepackage{subfigure}
\usepackage{graphicx}
\usepackage{algorithm}
\usepackage{algorithmic}
\usepackage{multirow}
\usepackage{xpatch}
\usepackage[capitalize,noabbrev]{cleveref}

\usepackage{caption}
\usepackage{hyperref}  
\title{Evading Visual Aphasia: Contrastive Adaptive Semantic Token Pruning for Vision-Language Models
}

\author{
  Jie Ma, Yihang Liu, Zhike Qiu, Jiayi Ji, Xiaoshuai Sun \\
  Xiamen University\\
  \texttt{\{jiema100,liuyihang\}@stu.xmu.edu.cn}, \texttt{zhikeqiu@outlook.com} \\
  \texttt{jjyxmu@gmail.com}, \texttt{xxsun@xmu.edu.cn} \\
}

\begin{document}
\maketitle

\begin{abstract}
Are low-attention visual tokens truly redundant in vision-language reasoning? Existing pruning methods often assume so, ranking visual tokens by shallow text-to-image attention and discarding low-scoring patches to accelerate LVLM inference. We show that this scalar criterion is unreliable for compositional reasoning: tokens ignored in early layers can later become essential for resolving secondary objects, spatial relations, and contextual cues. Premature pruning can therefore induce \textit{Visual Aphasia}, a failure mode in which the model loses visual grounding and falls back on language priors.
We introduce COAST (\textbf{CO}ntrastive \textbf{A}daptive \textbf{S}emantic \textbf{T}oken Pruning), a training-free pruning framework that casts compression as adaptive semantic routing. COAST uses native cross-modal attention to identify query-specific anchors and estimate contextual dispersion via attention entropy, then adapts the retention trade-off between semantic evidence and spatial context. It further uses a contrastive routing score to preserve both anchor-aligned evidence and complementary spatial context. 
Across seven benchmarks, COAST reduces visual tokens by 77.8\% and achieves a 2.15$\times$ latency speedup while retaining 98.64\% of the original average performance. Beyond a single backbone or compression setting, COAST consistently outperforms strong pruning baselines across token budgets and generalizes across multiple LVLM families, showing that adaptive semantic routing is a robust alternative to one-shot scalar pruning.

\end{abstract}


\section{Introduction}
\label{sec:intro}

Large Vision-Language Models (LVLMs)~\cite{gpt4,Gemini,llava,deepseekai2025deepseekv3technicalreport,qwenvl25} achieve strong multimodal reasoning by processing visual inputs and text instructions in a unified sequence~\cite{Flamingo,llava1}. To preserve fine-grained visual evidence, recent models encode images with dense patch representations~\cite{vit,winft,clip} or dynamic tiling strategies~\cite{llavanext,qwenvl25,internvl35advancingopensourcemultimodal}, producing sequences from hundreds to thousands of visual tokens. Although these tokens improve perceptual fidelity, many correspond to background regions or details irrelevant to a given instruction~\cite{Fastv,vispruner}. The resulting redundancy substantially increases inference cost, especially in decoder self-attention, whose complexity grows quadratically with sequence length~\cite{streamingllm,EfficientTransformers,AttentionisAllyouNeed,kwon2023vllm}.

\begin{figure}[!t]
\centering
\includegraphics[width=\linewidth]{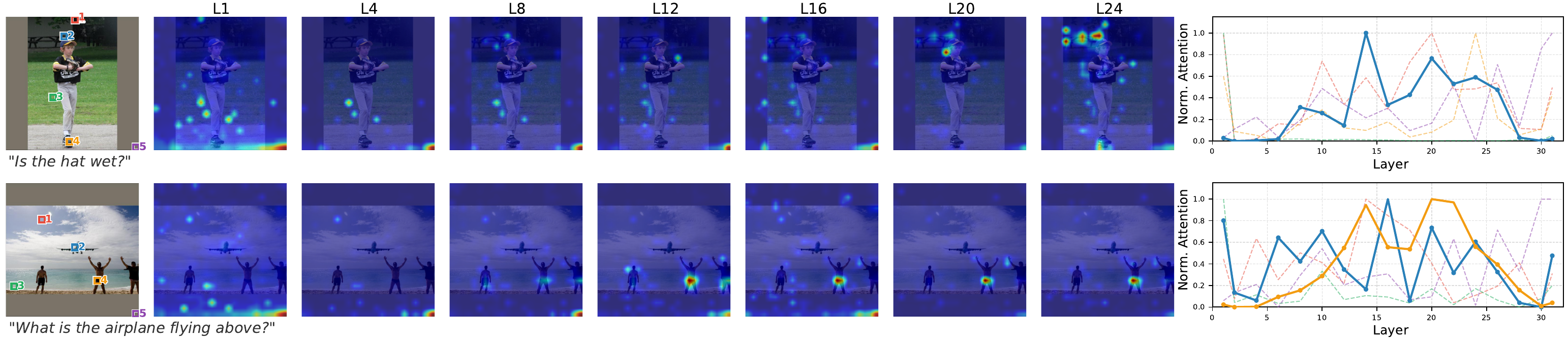}
\caption{
Are low-attention tokens truly redundant?
Attention trajectories reveal a limitation of early scalar pruning.
We track five visual tokens across transformer layers for two representative examples. Although these tokens receive low text-to-image attention at the shallow layer where FastV prunes, their attention increases in later layers. This indicates that early scalar attention can underestimate tokens that become relevant during deeper multimodal reasoning.
}
\label{fig:fig1_visual_aphasia}
\end{figure}

This cost has motivated visual token reduction methods~\cite{tome,Fastv,fasterVLM,HiPrune,holov,sparsevlm,vispruner,DivPrune} that compress the visual sequence before or during language decoding. Existing approaches~\cite{Fastv,sparsevlm,vispruner,holov} typically rely on early or layer-static importance estimates. Text-aware methods such as FastV~\cite{Fastv} prune tokens according to shallow-layer text-to-image attention, while spatial reduction methods such as LLaVA-PruMerge~\cite{llavaprumerge} merge patches based on local visual similarity. These strategies are effective for reducing computation, but they implicitly assume that token importance can be determined from an early, scalar signal. We find that this assumption is unreliable for compositional vision-language reasoning~\cite{gqa,CLEVR}. Tokens that receive low attention in shallow layers may later become important for resolving secondary objects, spatial relations, or contextual cues. As shown in~\cref{fig:fig1_visual_aphasia},  tokens prematurely discarded by early attention pruning often become crucial in deeper layers. This reveals that shallow saliency inherently underestimates contextually emerging visual evidence.

%

\begin{figure*}[ht] 
    \centering
    \includegraphics[width=\textwidth]{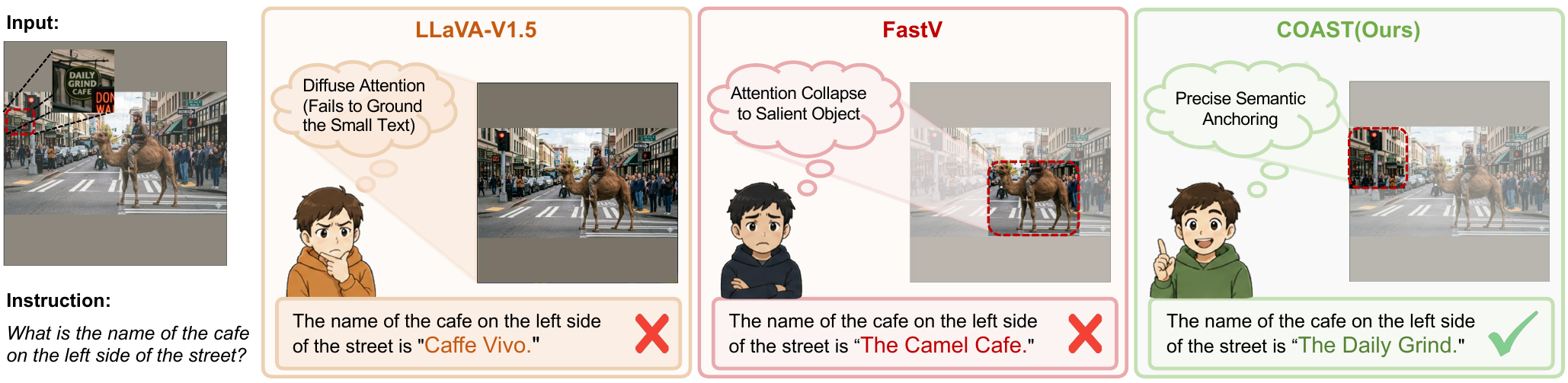}
    \caption{\textbf{A qualitative case of Visual Aphasia.}
    Given an image dominated by a salient object (the camel) and a question targeting a peripheral detail (the cafe sign on the left), Dense LLaVA-1.5 produces diffuse attention and hallucinates \textit{``Caffe Vivo''}, a name that does not appear in the image. FastV further amplifies this failure: its scalar attention pruning discards the small-text region in shallow layers and collapses the retained attention onto the most salient object, producing \textit{``The Camel Cafe''}---a textbook case of Visual Aphasia, where language priors fill in for missing visual evidence. In contrast, COAST preserves complementary spatial context through contrastive routing, maintains semantic anchoring on the \textit{``Daily Grind''} signage in deeper layers, and answers correctly.} 
    \label{fig:visual_aphasia_case}
\end{figure*}

We refer to this failure mode as \textbf{Visual Aphasia}: after premature pruning, the model retains the language instruction but loses part of the visual evidence needed to ground its reasoning. Unlike ordinary token redundancy~\cite{tome,Fastv,DynamicViT}, Visual Aphasia is not merely caused by removing visually uninformative patches; it arises when pruning disrupts the visual context required for later cross-modal coordination~\cite{agrawal2018dontjust,vqa2,WomenAlsoSnowboard}. The model may then answer using language priors or incomplete visual evidence. As highlighted in recent studies on multimodal reasoning~\cite{pope,HallusionBench}, LVLMs heavily suffer from an over-reliance on statistical language biases when visual grounding is weak~\cite{mitigating}, leading to errors in fine-grained recognition~\cite{tong2024eyeswideshut}, spatial reasoning~\cite{kamath2023whatsup}, and hallucination-sensitive tasks~\cite{bingo}.
\cref{fig:visual_aphasia_case} illustrates a textbook case of this failure mode. When asked about a small cafe sign, FastV's pruned attention collapses onto the most salient object (the camel), producing the hallucinated answer \textit{``The Camel Cafe''}, while COAST correctly anchors on the \textit{``DAILY GRIND''} signage.

To address this problem, we propose \textbf{COAST} (\textbf{CO}ntrastive \textbf{A}daptive \textbf{S}emantic \textbf{T}oken Pruning), a training-free pruning framework that treats visual compression as adaptive semantic routing. Rather than deciding token importance from a single shallow attention score, COAST separates two roles that are conflated in standard pruning: tokens that directly support the query semantics, and tokens that provide the spatial context needed to interpret them. Specifically, COAST reuses native cross-modal attention to identify query-specific visual anchors and to estimate scene-level contextual dispersion. It then uses attention entropy to adaptively allocate the retention budget between semantic evidence and spatial context, assigning more capacity to contextual preservation when the attention signal is dispersed.
Given this adaptive budget, candidate tokens are scored by contrasting their maximum similarity to semantic anchors with their average similarity to a contextual reference set. COAST retains candidates from both ends of this score distribution: high-scoring tokens preserve anchor-aligned evidence, while low-scoring tokens preserve complementary spatial context that would be missed by anchor-only pruning. In this way, COAST reduces visual redundancy without relying on brittle one-shot scalar decisions, maintaining visual grounding for later reasoning layers while improving inference efficiency.

Our main contributions are summarized as follows:
\begin{itemize}
    \item We identify \textit{Visual Aphasia}, a pruning-induced failure mode in which LVLMs lose visual grounding because shallow token importance fails to capture evidence needed in later reasoning layers.
    \item We introduce \textbf{COAST}, a training-free visual token routing method that combines anchor-based semantic selection, contextual preservation, and entropy-adaptive budget allocation.
    \item We conduct extensive evaluations on seven multimodal benchmarks, including MME, MMBench, AI2D, GQA, VizWiz, POPE, and ScienceQA-IMG. COAST reduces visual tokens by $77.8\%$ and achieves a $2.15\times$ latency speedup while retaining $98.64\%$ of the original average performance, outperforming strong pruning baselines across token budgets and transferring across multiple LVLM families.

\end{itemize}

\section{Related Work}

\subsection{Large Vision-Language Models}
Recent Large Vision-Language Models (LVLMs)~\cite{llava,qwen3technicalreport,qwenvl25,gpt4,deepseekai2025deepseekv3technicalreport} process visual inputs and text instructions within a unified sequence, enabling strong multimodal perception and reasoning. To preserve fine-grained visual evidence, modern LVLMs often represent high-resolution images with dense patch tokens or dynamic tiling, increasing the number of visual tokens from hundreds~\cite{llava} to thousands or more~\cite{llavanext,qwenvl25,qwen3technicalreport}. While this design improves visual fidelity, it also introduces substantial redundancy because many image regions are weakly related to a given instruction. The long visual sequence increases decoder self-attention cost and KV-cache memory consumption, motivating efficient visual token compression for resource-constrained and long-context LVLM deployment.

\subsection{Visual Token Reduction}
Existing visual token reduction methods can be broadly grouped into text-agnostic and text-aware approaches. \textit{Text-agnostic} methods~\cite{holov,llavaprumerge,fasterVLM,HiPrune,DivPrune} compress visual sequences using visual similarity, spatial merging, or clustering, often before the language decoder processes the instruction. These methods are efficient and architecture-friendly, but because they do not explicitly condition on the textual query, they may remove regions that are visually small yet important for the downstream task.
\textit{Text-aware} methods~\cite{Fastv,sparsevlm,vispruner,visionzip,PyramidDrop,btp} instead use cross-modal signals, such as text-to-image attention in the LLM decoder, to estimate token importance. This improves query relevance, but most methods still make pruning decisions from scalar saliency scores, often at early layers or under a fixed retention ratio. Such criteria can be brittle for compositional reasoning, where low-attention tokens in shallow layers may later support object disambiguation, spatial relation understanding, or contextual grounding. COAST differs from prior pruning methods by treating compression as adaptive semantic routing: it combines anchor-based semantic selection with complementary spatial context preservation, and uses attention entropy to adjust the retention trade-off across inputs.

\section{Methodology}
\label{sec:method}

\begin{figure*}[t]
\centering
\includegraphics[width=\textwidth]{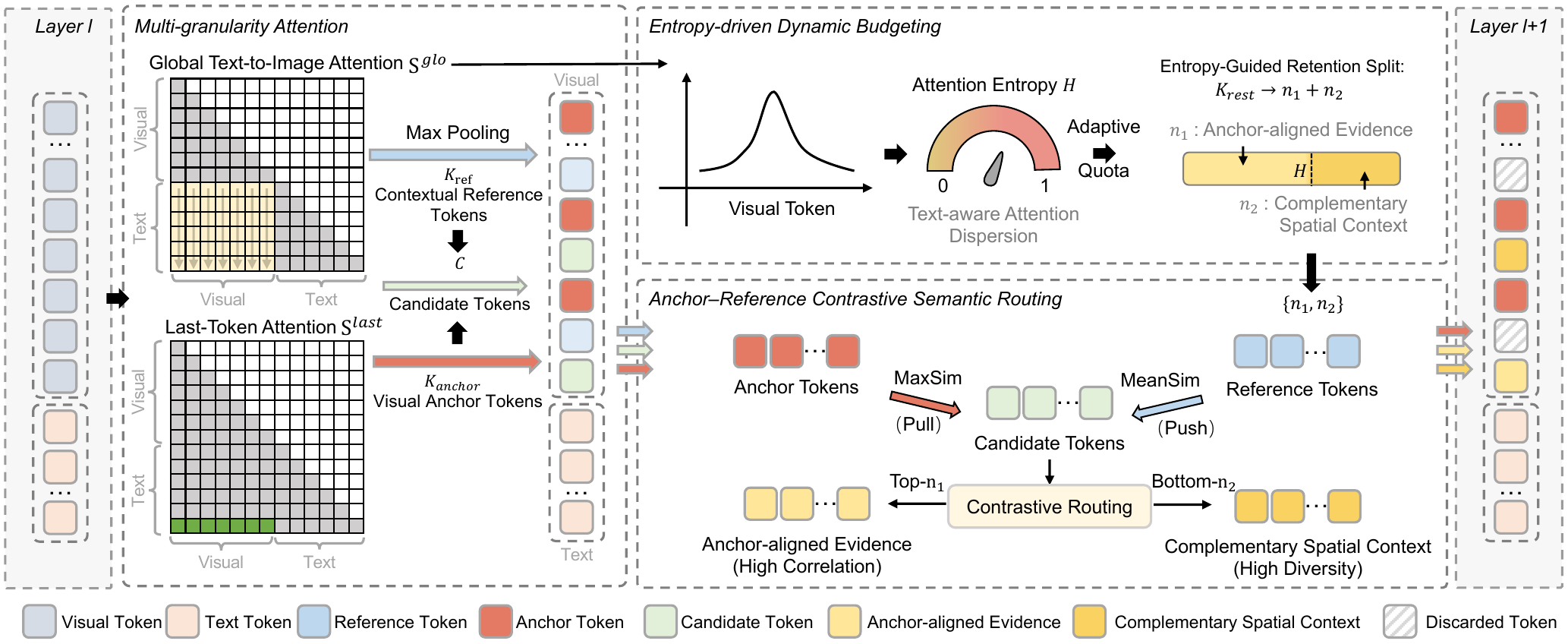}
\caption{
\textbf{Overview of COAST.}
At each scheduled pruning layer, COAST reuses cached cross-modal attention to select query-specific anchors from last-token attention $S^{last}$ and contextual reference tokens from global attention $S^{glo}$. Attention entropy guides the split of the remaining budget $K_{\mathrm{rest}}$ into anchor-aligned evidence ($n_1$) and complementary spatial context ($n_2$). Candidate tokens are scored by contrasting their MaxSim to anchors with their MeanSim to the reference set. COAST retains top-$n_1$ candidates as semantic evidence and bottom-$n_2$ candidates as complementary context, then sorts the retained tokens in their original order for the next layer.}
\label{fig:pipeline}
\end{figure*}

We first introduce the LVLM inference setup in~\cref{sec:preliminaries} and revisit the limitations of scalar attention pruning in~\cref{sec:rethinking}. Building on this analysis, we propose \textbf{COAST} (\textbf{CO}ntrastive \textbf{A}daptive \textbf{S}emantic \textbf{T}oken Pruning), a training-free framework that formulates visual compression as progressive semantic routing. As illustrated in~\cref{fig:pipeline}, COAST consists of two key components. \textbf{Entropy-Driven Dynamic Budgeting} (\cref{sec:budgeting}) reuses cached cross-modal attention to estimate attention dispersion and allocate the non-anchor retention budget between semantic evidence and complementary spatial context. \textbf{Contrastive Semantic Routing} (\cref{sec:routing}) then scores candidate tokens using their similarity to query-specific anchors and contextual reference tokens, retaining candidates from both ends of the score distribution for subsequent layers.

\subsection{Preliminaries}
\label{sec:preliminaries}

We consider a typical decoder-only LVLM~\cite{llava,llavanext,llavaOneVision,qwenvl25}, which processes visual and textual inputs in a unified token sequence. Given an image $I$ and a text instruction $T$, a vision encoder extracts visual features, which are projected into the language-model embedding space by a visual projector. This yields a visual token sequence $\mathbf{X}_v \in \mathbb{R}^{N_v \times d}$, where $N_v$ denotes the number of visual tokens and $d$ is the hidden dimension. The text instruction is embedded as $\mathbf{X}_t \in \mathbb{R}^{N_t \times d}$.
During the prefill stage, the concatenated sequence
    $\mathbf{X}^{(0)} = [\mathbf{X}_v, \mathbf{X}_t] \in \mathbb{R}^{(N_v + N_t) \times d}$
is fed into an $L$-layer Transformer decoder. At layer $l$, multi-head self-attention updates the hidden states as
\begin{equation}
    \mathbf{X}^{(l)} = \mathrm{MHSA}^{(l)}(\mathbf{X}^{(l-1)}) + \mathbf{X}^{(l-1)} ,
\end{equation}
where feed-forward and normalization operations are omitted for simplicity. The self-attention cost scales quadratically with the sequence length, i.e.,
$\mathcal{O}((N_v+N_t)^2 d)$, and the retained visual tokens also contribute to KV-cache memory across decoder layers.

COAST focuses on reducing the number of visual tokens during prefill. At selected pruning layers, it retains all text tokens and removes a subset of visual tokens before passing the sequence to subsequent layers. Let $\mathcal{P}$ denote the set of pruning layers. For a pruning layer $l \in \mathcal{P}$, we use $N_v^{(l)}$ and $K_v^{(l)}$ to denote the number of visual tokens available before and after pruning at layer $l$, respectively, where $K_v^{(l)} < N_v^{(l)}$. The goal is to reduce visual-token computation across subsequent layers while preserving the visual evidence required for cross-modal reasoning.

\subsection{Rethinking Attention-Based Pruning and Visual Aphasia}
\label{sec:rethinking}

Many visual token pruning methods reduce inference cost by ranking visual tokens with a scalar saliency score and retaining only the top-scoring subset. For example, attention-based methods such as FastV~\cite{Fastv} use shallow text-to-image attention as the importance estimate. At a pruning layer $l$, let $\mathbf{X}_v^{(l)} \in \mathbb{R}^{N_v^{(l)} \times d}$ denote the current visual token sequence before pruning, where $N_v^{(l)}$ is the number of available visual tokens. Let $\mathbf{s}^{(l)} \in \mathbb{R}^{N_v^{(l)}}$ be a scalar importance score over these tokens, and let $K_v^{(l)} < N_v^{(l)}$ be the target number of visual tokens retained after pruning. A standard scalar top-$K$ pruning rule can be written as
\begin{equation}
    \mathcal{I}_{\mathrm{top}}^{(l)}
    =
    \operatorname{TopK}\!\left(\mathbf{s}^{(l)}, K_v^{(l)}\right),
    \qquad
    \widetilde{\mathbf{X}}_v^{(l)}
    =
    \mathbf{X}_v^{(l)}[\mathcal{I}_{\mathrm{top}}^{(l)}],
\end{equation}
where $\operatorname{TopK}(\mathbf{s}^{(l)}, K_v^{(l)})$ returns the indices of the $K_v^{(l)}$ largest entries in $\mathbf{s}^{(l)}$, and $\mathbf{X}_v^{(l)}[\mathcal{I}]$ denotes gathering tokens according to index set $\mathcal{I}$ while preserving their original order. This rule is simple and efficient, but it makes pruning decisions along a single scalar axis. We argue that this design has two limitations for compositional vision-language reasoning.

\textbf{Conflating semantic evidence and spatial context.}
Scalar top-$K$ pruning ranks all visual tokens with one importance score, without distinguishing the roles that different tokens play in reasoning. Tokens that directly match the query semantics and tokens that provide surrounding spatial context compete under the same ranking criterion. As a result, the retained set can become overly concentrated on the most salient regions while discarding contextual evidence needed to interpret object relations, spatial layouts, or scene-level cues. This motivates a budget allocation strategy that explicitly balances semantic evidence and complementary spatial context (\cref{sec:budgeting}).

\textbf{Instability of shallow saliency.}
Low attention in early layers does not necessarily imply semantic redundancy. In compositional tasks, tokens corresponding to secondary objects, spatial boundaries, or peripheral cues may receive little attention before deeper cross-modal interactions are formed. Once such tokens are removed, later layers cannot recover the missing visual evidence (a phenomenon we statistically substantiate across multiple benchmarks in \cref{app:recovery_curve}). We refer to the resulting failure mode as \textit{Visual Aphasia}: the model retains the language instruction, but its reasoning becomes weakly grounded in the image and more dependent on language priors.

These observations suggest that visual compression should not be reduced to one-shot scalar truncation. A more reliable pruning rule should preserve both query-aligned evidence and complementary context, while adapting this trade-off to the attention structure of each input. COAST achieves this via entropy-driven dynamic budgeting (\cref{sec:budgeting}) and contrastive semantic routing (\cref{sec:routing}).

\begin{figure}[t]
\centering
\includegraphics[width=\textwidth]{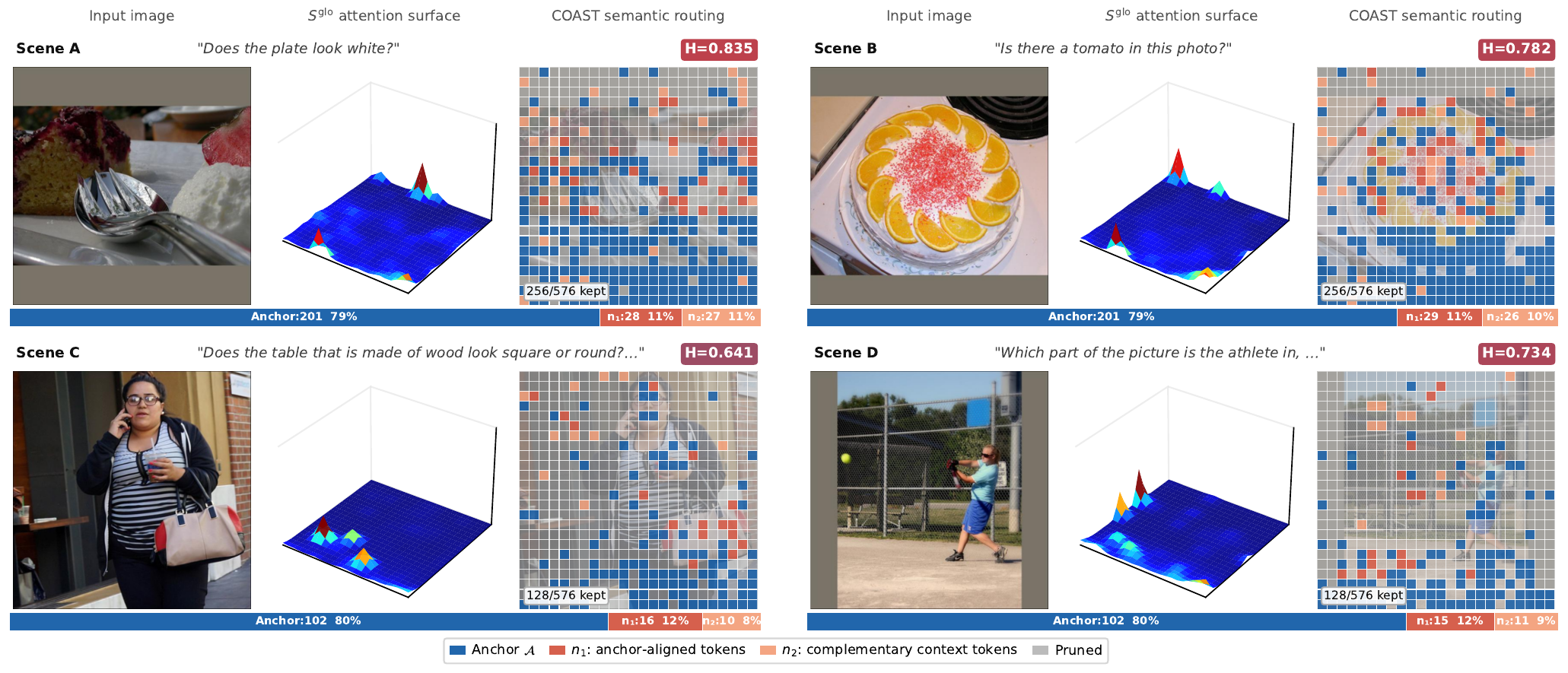}
\caption{Entropy-driven dynamic budget allocation across diverse scenes. For each scene, we visualize the input image, the $S^{glo}$ attention surface, and the COAST semantic routing result. The attention surface reveals the distribution of cross-modal attention: sharper peaks indicate more focused semantics (lower $H$), while flatter landscapes reflect dispersed attention (higher $H$). As H increases from Scene C ($H=0.641$) to Scene A ($H=0.835$), COAST adaptively shifts the retention budget toward complementary context tokens ($n_2: 8\%\to11\%$), while the anchor ratio remains stable (e.g., $\sim80\%$), illustrating how the entropy-driven split adapts the retained-token composition to the attention distribution.}
\label{fig:ana2}
\end{figure}

\subsection{Entropy-Driven Dynamic Budgeting}
\label{sec:budgeting}

A fixed pruning ratio specifies how many visual tokens to keep, but it does not specify what kinds of tokens should occupy the retained set. For vision-language reasoning, this composition matters: some inputs contain a compact query target, while others require broader contextual coverage to resolve spatial relations or secondary objects. COAST therefore uses cross-modal attention entropy to adaptively split the retention budget between semantic evidence and complementary spatial context.

For a pruning layer $l$, let $N_v^{(l)}$ be the number of visual tokens before pruning and $K_v^{(l)}$ be the target number retained after pruning. COAST reuses the attention probability matrix $\mathbf{A}^{(l-1)}$ from the preceding layer and averages it over heads. Let $\mathbf{A}_{T\to V}^{(l-1)} \in \mathbb{R}^{N_t \times N_v^{(l)}}$ denote the text-to-image attention submatrix over the current visual tokens. We compute a global attention score by max-pooling over text tokens:
\begin{equation}
    S^{glo}_j = \max_{i=1}^{N_t} \mathbf{A}_{T\to V}^{(l-1)}(i,j),
    \quad j=1,\ldots,N_v^{(l)} .
\end{equation}
This captures whether each visual token is strongly attended by any part of the instruction. We then normalize $S^{glo}$ into $p_j = S^{glo}_j / \sum_m S^{glo}_m$ and compute the normalized entropy
\begin{equation}
    H^{(l)}
    =
    -\frac{1}{\log N_v^{(l)}}
    \sum_{j=1}^{N_v^{(l)}} p_j \log p_j .
\end{equation}
A low entropy indicates that attention is concentrated on a small set of visual regions, whereas a high entropy indicates more dispersed cross-modal evidence.

COAST first reserves query-specific anchor tokens $K_A^{(l)}$, selected by last-token attention $S^{last}$. The remaining budget
\begin{equation}
    K_{\mathrm{rest}}^{(l)} = K_v^{(l)} - K_A^{(l)}
\end{equation}
is then split into semantic and contextual portions:
\begin{align}
    n_2^{(l)}
    &=
    \left\lfloor
    K_{\mathrm{rest}}^{(l)}
    \left(
    \alpha_{\min}
    +
    (\alpha_{\max}-\alpha_{\min}) H^{(l)}
    \right)
    \right\rfloor,\\
    n_1^{(l)}
    &=
    K_{\mathrm{rest}}^{(l)} - n_2^{(l)} .
\end{align}
Here, $n_1^{(l)}$ is the number of additional anchor-aligned candidates and $n_2^{(l)}$ is the number of complementary context candidates. The bounds $\alpha_{\min}$ and $\alpha_{\max}$ control the minimum and maximum fraction of the remaining budget assigned to contextual preservation. Thus, COAST does not change the total target budget $K_v^{(l)}$; instead, it adapts the composition of the retained set. Concentrated attention favors semantic evidence, while dispersed attention assigns more capacity to spatial context. As visualized in~\cref{fig:ana2}, this entropy-guided split assigns more budget to complementary spatial context when global attention is diffuse, and favors anchor-aligned evidence when attention is concentrated.

\subsection{Anchor--Reference Contrastive Semantic Routing}
\label{sec:routing}

Given the budget split from \cref{sec:budgeting}, COAST determines which non-anchor tokens should be retained. The goal is to preserve two complementary types of visual evidence: tokens that align with query-specific anchors and tokens that provide spatial context missed by anchor-only selection. COAST therefore scores candidate tokens by their relative similarity to semantic anchors and contextual reference tokens, and retains candidates from both ends of the resulting score distribution.

At pruning layer $l$, the semantic anchor set $\mathcal{A}^{(l)}$ contains the $K_A^{(l)}$ tokens selected by last-token attention. COAST also constructs a contextual reference set $\mathcal{R}^{(l)}$ by selecting the $K_R^{(l)}$ lowest-scoring tokens under the global attention score $S^{glo}$. The reference set is not treated as semantic evidence; instead, it provides a low-salience reference for identifying candidate tokens outside the anchor-aligned region. The remaining non-anchor visual tokens form the candidate pool $\mathcal{C}^{(l)}$.

For each candidate token $\mathbf{c}_i \in \mathcal{C}^{(l)}$, COAST first computes its maximum cosine similarity to the anchor set:
\begin{equation}
    \mathrm{Sim}_{A}(\mathbf{c}_i)
    =
    \max_{\mathbf{a}_j \in \mathcal{A}^{(l)}}
    \frac{\mathbf{c}_i^\top \mathbf{a}_j}
    {\|\mathbf{c}_i\|_2 \|\mathbf{a}_j\|_2}.
\end{equation}
The max operation allows a candidate token to match any individual anchor, which is useful when the anchor set covers multiple objects or visual attributes. COAST then computes the candidate token's average similarity to the contextual reference set:
\begin{equation}
    \mathrm{Sim}_{R}(\mathbf{c}_i)
    =
    \frac{1}{|\mathcal{R}^{(l)}|}
    \sum_{\mathbf{r}_k \in \mathcal{R}^{(l)}}
    \frac{\mathbf{c}_i^\top \mathbf{r}_k}
    {\|\mathbf{c}_i\|_2 \|\mathbf{r}_k\|_2}.
\end{equation}
The resulting anchor--reference score is
\begin{equation}
    \mathrm{Score}(\mathbf{c}_i)
    =
    \mathrm{Sim}_{A}(\mathbf{c}_i)
    -
    \mathrm{Sim}_{R}(\mathbf{c}_i).
\end{equation}

COAST applies a two-tail retention rule to this score distribution. The top-$n_1^{(l)}$ candidates are retained as anchor-aligned evidence because this subset is close to at least one semantic anchor relative to the contextual reference. The bottom-$n_2^{(l)}$ candidates are retained as complementary spatial context because this subset lies outside the anchor-aligned region and helps preserve visual coverage. This two-tail selection prevents the compressed visual sequence from being dominated only by the most salient query regions.
Finally, COAST combines the anchor tokens, top-$n_1^{(l)}$ candidates, and bottom-$n_2^{(l)}$ candidates:
\begin{equation}
    \mathcal{I}_{\mathrm{keep}}^{(l)}
    =
    \mathcal{I}_{\mathrm{text}}
    \cup
    \mathrm{Idx}(\mathcal{A}^{(l)})
    \cup
    \mathrm{Idx}(\mathcal{C}^{(l)}_{\mathrm{top}\text{-}n_1})
    \cup
    \mathrm{Idx}(\mathcal{C}^{(l)}_{\mathrm{bottom}\text{-}n_2}) .
\end{equation}
Here, $\mathrm{Idx}(\cdot)$ returns the original sequence indices of the selected visual tokens. COAST sorts the retained indices in their original order before truncating the hidden states, position ids, and attention mask. Ordered recomposition allows subsequent decoder layers to process a shorter sequence while preserving the original token order. Additional qualitative visualizations are provided in the \cref{sec:appendix_visualization} to illustrate how COAST preserves complementary spatial context during token routing.

\paragraph{Complexity and FLOPs.}
 COAST introduces a lightweight routing overhead per pruning layer. For candidate set $\mathcal{C}^{(l)}$, anchor set $\mathcal{A}^{(l)}$, and reference set $\mathcal{R}^{(l)}$, the additional similarity computation costs 
\begin{equation}
    \mathcal{O}\!\left(
    |\mathcal{C}^{(l)}|
    \left(
    |\mathcal{A}^{(l)}| + |\mathcal{R}^{(l)}|
    \right)d
    \right),
\end{equation}
which is small because $\mathcal{A}^{(l)}$ and $\mathcal{R}^{(l)}$ are compact subsets.
The savings come from reducing the sequence length processed by subsequent decoder layers. Let $n_l$ denote the total sequence length after pruning at layer $l$, including text and retained visual tokens. For a decoder layer with hidden size $d$ and FFN intermediate size $m$, the dominant FLOPs are approximated by
\begin{equation}
    \mathrm{FLOPs}_{\mathrm{layer}}(n_l)
    \approx
    8 n_l d^2
    +
    4 n_l^2 d
    +
    6 n_l d m ,
\end{equation}
where the first two terms correspond to multi-head self-attention and the last term corresponds to the feed-forward network. By progressively reducing visual tokens, COAST lowers both the quadratic attention term and the linear projection/FFN terms in all subsequent layers. In experiments, we report FLOPs using this layer-wise estimate with the actual retained sequence lengths.
\section{Experiment}
\label{sec:experiment}
\subsection{Experimental setup}

\paragraph{Models and Baselines.}
To comprehensively evaluate the proposed COAST paradigm, we integrate it into representative LVLMs, including LLaVA-v1.5 (7B/13B)~\cite{llava}, LLaVA-NEXT-7B~\cite{llavanext}, LLaVA-OneVision-0.5B~\cite{llavaOneVision}, and Qwen2.5-VL-7B-Instruct~\cite{qwenvl25}. We systematically compare COAST against recent training-free visual token reduction baselines under a unified setting: FastV~\cite{Fastv}, PDrop~\cite{PyramidDrop}, SparseVLM~\cite{sparsevlm}, V$^2$Drop~\cite{v2drop}, and MMTOK~\cite{mmtok}.

\paragraph{Benchmarks and Evaluation Setup.}
We evaluate across diverse multimodal benchmarks: MME~\cite{mme}, MMBench-EN~\cite{mmb}, AI2D~\cite{AI2D}, GQA~\cite{gqa}, VizWiz~\cite{VizWiz}, POPE~\cite{pope}, and ScienceQA-IMG~\cite{sqa}. Following the LMMs-Eval framework~\cite{lmms_eval2024} for strict reproducibility, we assess all methods by balancing accuracy with computational efficiency, measured by estimated FLOPs and inference latency. Hardware configurations and detailed pruning schedules are deferred to \cref{app:implementation}.

\subsection{Main results}

Table~\ref{tab:main_results} presents a comparison between COAST and several representative token compression methods on LLaVA-1.5-7B under different visual token budgets
(128, 192, and 288), 
reporting both performance and efficiency metrics. 
COAST achieves the best or near-best average performance across all settings while maintaining competitive computational cost, demonstrating a favorable performance–efficiency trade-off.
Specifically, COAST exhibits the smallest performance degradation across different compression ratios. Notably, under the 192-token setting, COAST incurs only a $0.21\%$ performance drop, substantially lower than SparseVLM ($1.61\%$) and V$^2$Drop ($1.47\%$).
Even under more aggressive compression, COAST effectively preserves model performance, indicating a stronger capability in retaining critical visual information.
In addition, COAST consistently achieves leading results on multiple benchmarks, such as MME and GQA, further demonstrating its robustness and generalization ability.
Crucially, despite operating under comparable FLOPS, COAST achieves lower latency than recent methods (e.g., $89.93$ ms vs $113.78$ ms for V$^2$Drop under the 192-token setting) while preserving stronger overall performance.

\begin{table}[t]
\centering
\renewcommand{\arraystretch}{0.6}
\footnotesize
\caption{Performance and efficiency comparison across different token budgets. The relative performance drop compared to the Original model is shown in parentheses, with the \textcolor{green!60!black}{\textbf{smallest drop}}.}
\label{tab:main_results}
\resizebox{\textwidth}{!}{
\begin{tabular}{l *{8}{c} cc} 
\toprule
\multirow{2}{*}{\textbf{Method}} & \multicolumn{8}{c}{\textbf{Performance Benchmarks}} & \multicolumn{2}{c}{\textbf{Efficiency}} \\
\cmidrule(lr){2-9} \cmidrule(l){10-11}
& \textbf{MME} & \textbf{MMB$_{en}$} & \textbf{AI2D} & \textbf{GQA} & \textbf{VizWiz} & \textbf{POPE} & \textbf{SQA} & \textbf{Avg. (\%)} & \textbf{TFLOPS} & \textbf{Latency} \\
\midrule
Original(LLaVA-v1.5-7B) & 1866.10 & 64.09 & 55.21 & 61.94 & 54.01 & 93.78 & 69.91 & 100.00 & 8.54 & 167.50 \\

\midrule 
\rowcolor{gray!15} \multicolumn{11}{c}{\textit{Retain 128 Tokens}} \\
\midrule
FastV \textcolor{blue}{(ECCV24)}     & 1735.09 & 61.68 & 53.59 & 54.60 & 54.99 & 96.62 & 68.72 & 96.80 ($\downarrow$ 3.20) & 2.62 & 74.28 \\
PDrop \textcolor{blue}{(CVPR25)}     & 1666.61 & 61.60 & 53.43 & 55.72 & 52.91 & 94.24 & 69.76 & 95.77 ($\downarrow$ 4.23) & 2.68 & 67.59 \\
SparseVLM \textcolor{blue}{(ICML25)} & 1755.64 & 63.66 & 54.57 & 58.51 & 53.71 & 94.08 & 68.67 & 97.81 ($\downarrow$ 2.19) & 2.90 & 68.73 \\
V$^2$Drop \textcolor{blue}{(CVPR26)} & 1712.59 & 59.28 & 53.63 & 56.19 & 55.35 & 94.72 & 68.37 & 96.20 ($\downarrow$ 3.80) & 2.59 & 92.80 \\
MMTOK \textcolor{blue}{(ICLR26)}     & 1747.78 & 62.46 & 54.11 & 59.48 & 55.15 & 91.81 & 65.53 & 96.99 ($\downarrow$ 3.01) & 2.42 & 80.47 \\
\rowcolor{gray!10} 
\textbf{Ours}               & 1864.48 & 63.57 & 54.57 & 58.80 & 52.86 & 95.04 & 68.77 & 98.64 (\textcolor{green!60!black}{\textbf{$\downarrow$ 1.36}}) & 2.60 & 77.94 \\

\midrule
\rowcolor{gray!15} \multicolumn{11}{c}{\textit{Retain 192 Tokens}} \\
\midrule
FastV \textcolor{blue}{(ECCV24)}     & 1792.64 & 62.97 & 53.95 & 57.77 & 54.87 & 95.84 & 69.31 & 98.32 ($\downarrow$ 1.68) & 3.44 & 84.22 \\
PDrop \textcolor{blue}{(CVPR25)}     & 1784.03 & 62.63 & 53.30 & 57.35 & 52.33 & 94.88 & 69.26 & 97.08 ($\downarrow$ 2.92) & 3.46 & 81.94 \\
SparseVLM \textcolor{blue}{(ICML25)} & 1780.50 & 63.75 & 54.53 & 59.42 & 54.40 & 93.91 & 68.72 & 98.39 ($\downarrow$ 1.61) & 3.86 & 92.07 \\
V$^2$Drop \textcolor{blue}{(CVPR26)} & 1818.75 & 62.37 & 54.73 & 58.60 & 55.58 & 93.42 & 68.96 & 98.53 ($\downarrow$ 1.47) & 3.42 & 113.78 \\
MMTOK \textcolor{blue}{(ICLR26)}     & 1785.06 & 63.66 & 54.89 & 60.22 & 54.69 & 91.90 & 65.39 & 97.77 ($\downarrow$ 2.23) & 3.25 & 94.62 \\
\rowcolor{gray!10} 
\textbf{Ours}               & 1898.94 & 63.92 & 55.34 & 61.03 & 53.77 & 93.77 & 69.01 & 99.79 (\textcolor{green!60!black}{\textbf{$\downarrow$ 0.21}}) & 3.44 & 89.93 \\

\midrule
\rowcolor{gray!15} \multicolumn{11}{c}{\textit{Retain 288 Tokens}} \\
\midrule
FastV \textcolor{blue}{(ECCV24)}     & 1838.94 & 63.75 & 55.12 & 59.90 & 54.19 & 95.33 & 68.67 & 99.25 ($\downarrow$ 0.75) & 4.70 & 125.80 \\
PDrop \textcolor{blue}{(CVPR25)}     & 1847.65 & 64.35 & 54.92 & 60.60 & 53.42 & 93.94 & 69.31 & 99.28 ($\downarrow$ 0.72) & 4.72 & 105.96 \\
SparseVLM \textcolor{blue}{(ICML25)} & 1849.29 & 64.08 & 54.92 & 61.23 & 53.45 & 93.47 & 68.42 & 99.13 ($\downarrow$ 0.87) & 5.02 & 109.12 \\
V$^2$Drop \textcolor{blue}{(CVPR26)} & 1822.67 & 64.09 & 55.38 & 60.77 & 55.47 & 93.35 & 69.06 & 99.59 ($\downarrow$ 0.41) & 4.68 & 136.40 \\
MMTOK \textcolor{blue}{(ICLR26)}     & 1792.00 & 64.09 & 54.92 & 61.20 & 54.12 & 96.11 & 65.27 & 98.62 ($\downarrow$ 1.38) & 4.51 & 125.75 \\
\rowcolor{gray!10} 
\textbf{Ours}               & 1856.35 & 64.18 & 54.99 & 61.43 & 54.38 & 93.32 & 69.11 & 99.63 (\textcolor{green!60!black}{\textbf{$\downarrow$ 0.37}}) & 4.68 & 104.45 \\
\bottomrule
\end{tabular}
}
\end{table}

\begin{figure}[t] 
  \centering
    \includegraphics[width=\linewidth]{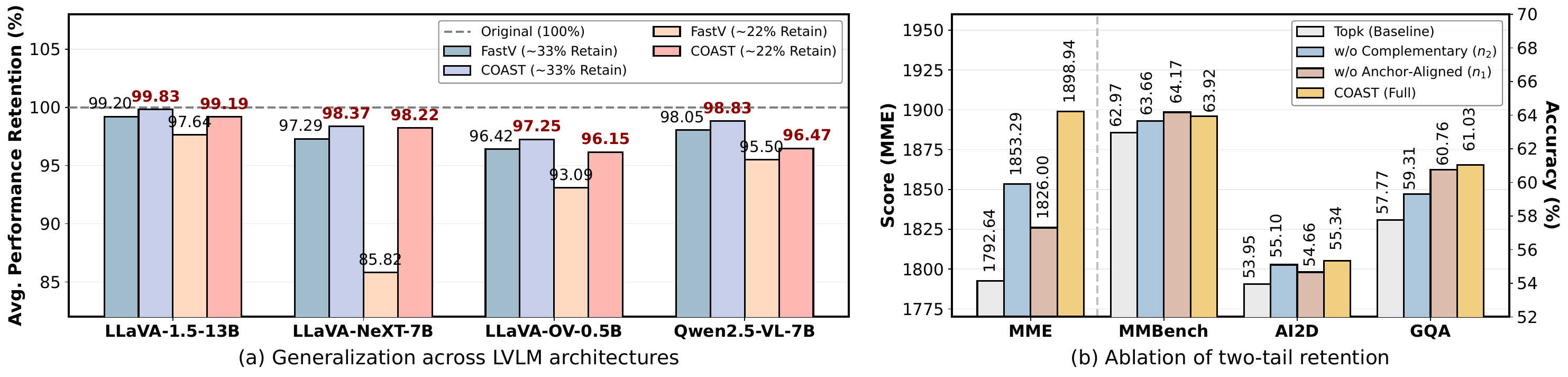}
\caption{Generalization and ablation analysis.
(a) Average performance retention relative to the original unpruned model (100\%, dashed line) across four LVLM backbones. COAST consistently outperforms FastV under both moderate ($\sim$33\%) and aggressive ($\sim$22\%) visual-token retention budgets, with larger gains under stronger compression.
(b) Ablation of two-tail retention under the same retention ratio. Anchor-aligned evidence ($n_1$) and complementary spatial context ($n_2$) provide complementary benefits: removing either branch changes the performance profile, while the full COAST achieves the strongest overall trade-off.}
\label{fig:generalization_ablation}
\end{figure}

\paragraph{Generalizability Across Architectures.}
To evaluate whether COAST generalizes beyond a single LVLM backbone, we apply it to four model families and scales: LLaVA-1.5-13B, LLaVA-NeXT-7B, LLaVA-OV-0.5B, and Qwen2.5-VL-7B. As shown in~\Cref{fig:generalization_ablation}(a), COAST outperforms the FastV baseline under both moderate ($\sim$33\%) and aggressive ($\sim$22\%) visual-token retention budgets. Under the moderate budget, COAST preserves most of the original model performance, reaching $99.83\%$ retention on LLaVA-1.5-13B and $98.83\%$ on Qwen2.5-VL-7B. The advantage becomes more pronounced under aggressive compression. For example, on LLaVA-NeXT-7B, FastV drops to $85.82\%$ retention at the $\sim$22\% budget, whereas COAST maintains $98.22\%$. Similar trends are observed on LLaVA-OV-0.5B and Qwen2.5-VL-7B. These results indicate that anchor--reference routing is not tied to a specific LVLM architecture and can preserve visual evidence across different model scales and visual-tokenization strategies. A full benchmark breakdown is provided in~\cref{appendix:appendix_generalization}.

\subsection{Ablation Studies}
\label{sec:ablation}

\paragraph{Two-tail retention.}

We ablate the two retained candidate groups in COAST under the same visual-token retention ratio. As shown in~\Cref{fig:generalization_ablation}(b), we compare COAST with a Top-K scalar pruning baseline and two single-path variants: removing complementary context (w/o $n_2$) and removing anchor-aligned candidates (w/o $n_1$). The anchor-only variant (w/o $n_2$) improves MME from $1792.64$ to $1853.29$ and AI2D from $53.95$ to $55.10$, indicating that anchor-aligned candidates preserve query-relevant evidence. Conversely, the context-only variant (w/o $n_1$) performs better on MMBench ($64.17$) and GQA ($60.76$), suggesting that complementary spatial context benefits scene-level reasoning. By combining both groups, COAST achieves the best results on MME ($1898.94$), AI2D ($55.34$), and GQA ($61.03$), while remaining competitive on MMBench ($63.92$). Overall, the two groups provide complementary benefits under aggressive compression, validating the effectiveness of the dual-path retention design.
Additional qualitative visualizations of token routing are provided in \cref{fig:routing_visualization}.
\begin{table}[!t]
\centering
\renewcommand{\arraystretch}{0.55}

\begin{minipage}[t]{.49\textwidth} 
\centering
\caption{\textbf{Ablation on entropy-driven dynamic budgeting.} Fixed ratios assign the same fraction of $K_{\mathrm{rest}}$ to complementary spatial context for every input. Entropy-guided allocation adapts this fraction using attention entropy and achieves the best overall performance across benchmarks.}
\label{tab:budget_ablation}
\resizebox{\linewidth}{!}{ 
\begin{tabular}{lcccc}
\toprule
\textbf{Budget split} & \textbf{MME} & \textbf{MMB$_{en}$} & \textbf{AI2D} & \textbf{GQA} \\
\midrule
Fixed 0.2 & 1861.35 & 63.49 & \textbf{55.41} & 60.33 \\
Fixed 0.4 & 1868.84 & 63.57 & 55.21 & 60.88 \\
Fixed 0.5 & 1889.10 & 63.83 & 55.31 & 61.02 \\
\midrule
\rowcolor{gray!10} 
\textbf{Entropy-guided} & \textbf{1898.94} & \textbf{63.92} & 55.34 & \textbf{61.03} \\
\bottomrule
\end{tabular}
}
\end{minipage}
\hfill 
%
\begin{minipage}[t]{.49\textwidth}
\centering
\caption{\textbf{Ablation on anchor--reference contrastive routing.} All variants use the same pruning schedule and retention budget. Scalar uses attention scores for candidate selection; MaxSim uses only similarity to semantic anchors; Contrastive uses the full anchor--reference score $\mathrm{Sim}_A-\mathrm{Sim}_R$.}
\label{tab:routing_ablation}
\resizebox{\linewidth}{!}{
\begin{tabular}{lcccc}
\toprule
\textbf{Routing score} & \textbf{MME} & \textbf{MMB$_{en}$} & \textbf{AI2D} & \textbf{GQA} \\
\midrule
Scalar attention & 1792.64 & 62.97 & 53.95 & 57.77 \\
MaxSim anchor-only & 1860.05 & 63.66 & 55.21 & 61.00 \\
\midrule
\rowcolor{gray!10}
\textbf{Contrastive} & \textbf{1898.94} & \textbf{63.92} & \textbf{55.34} & \textbf{61.03} \\
\bottomrule
\end{tabular}
}
\end{minipage}
\end{table}

\paragraph{Entropy-Driven Dynamic Budgeting.}
We study whether the semantic-contextual budget should be fixed or adapted per input. Table~\ref{tab:budget_ablation} compares entropy-guided allocation with fixed context ratios, where the ratio denotes the fraction of $K_{\mathrm{rest}}$ assigned to complementary spatial context for every sample. Entropy-guided budgeting achieves the best results on MME, MMBench, and GQA, reaching $1898.94$, $63.92$, and $61.03$, respectively. Compared with the strongest fixed ratio ($0.5$), it further improves MME by $9.84$ points and gives small gains on MMBench and GQA. On AI2D, the fixed $0.2$ ratio is marginally higher ($55.41$ vs. $55.34$), suggesting that diagram-oriented tasks may prefer a more semantic-heavy allocation. Overall, the entropy-guided split provides the most robust cross-benchmark trade-off without requiring a manually tuned global context ratio.

\paragraph{Anchor--Reference Contrastive Routing.}

We evaluate the scoring function used to route candidate tokens. Table~\ref{tab:routing_ablation} compares three variants under the same pruning schedule and retention budget: scalar attention scoring, anchor-only MaxSim scoring, and the full anchor--reference contrastive score. Replacing scalar attention with MaxSim substantially improves performance, increasing MME from $1792.64$ to $1860.05$ and GQA from $57.77$ to $61.00$. This indicates that feature-space similarity to query-specific anchors provides a stronger signal than shallow scalar attention. Adding the contextual reference term further improves all benchmarks, reaching $1898.94$ on MME, $63.92$ on MMBench, $55.34$ on AI2D, and $61.03$ on GQA. These results show that contrastive anchor--reference scoring better separates query-aligned evidence from complementary context under aggressive compression.

\section{Conclusion}

We presented COAST, a training-free visual token routing framework for LVLM inference. Motivated by the observation that low-attention visual tokens can become important in later reasoning layers, we identified \textit{Visual Aphasia} as a pruning-induced failure mode in which aggressive early truncation weakens visual grounding and increases reliance on language priors. COAST replaces one-shot scalar pruning with adaptive semantic routing: it uses attention entropy to balance anchor-aligned evidence and complementary spatial context, and applies anchor--reference contrastive scoring to retain candidates from both ends of the score distribution. Experiments and ablations show that preserving both semantic anchors and spatial context improves the accuracy-efficiency trade-off under aggressive compression, effectively mitigating Visual Aphasia across diverse LVLM architectures.

\bibliographystyle{unsrt}  
\bibliography{references}  

\newpage
\appendix
\section{Appendix}
\label{app:appendix}

\subsection{Additional Implementation Details}
\label{app:implementation}

\subsubsection{Model and Inference Setup}

COAST is a training-free visual token routing method and does not require finetuning the LVLM or adding auxiliary modules. Unless otherwise specified, we apply COAST during the prefill stage of the language decoder and keep all text tokens throughout inference. Only visual tokens are pruned. The visual token span is determined by the input template of the underlying LVLM.
All experiments are conducted on LLaVA-v1.5, LLaVA-NEXT, LLaVA-OneVision,and Qwen2.5-VL.
We implement the proposed COAST framework using PyTorch (v2.1.2) and Transformers (v4.37.2). All experiments are conducted on eight NVIDIA RTX 3090 GPUs. For a fair and reproducible comparison, all benchmark evaluations are performed using the lmm-eval(v0.4.1) framework~\cite{lmms_eval2024} with the same target visual-token budget.

\subsubsection{COAST Hyperparameters}

COAST uses a fixed hyperparameter configuration across all benchmarks; we do not tune the parameters separately for individual datasets. The hyperparameters mainly control the pruning schedule, anchor/reference selection, and entropy-guided split. The pruning schedule specifies the target number of visual tokens retained after each scheduled pruning layer. The anchor and reference ratios determine the candidate sets used by anchor--reference routing, while $\alpha_{\min}$ and $\alpha_{\max}$ bound the fraction of the remaining budget assigned to complementary spatial context.

\begin{table}[h]
\centering

\caption{COAST hyperparameters used in the main experiments.
We report the hyperparameters shared across pruning budgets. The pruning schedule and target visual-token counts are adjusted according to the specified retention ratio.}
\label{tab:app_hyperparams}
\resizebox{0.4\textwidth}{!}{
\begin{tabular}{lc}
\toprule
Hyperparameter & Value \\
\midrule
Anchor ratio $\rho_A$ & $0.30$ \\
Reference ratio $\rho_R$ & $0.05$ \\
Maximum anchor budget ratio $\eta$ & $0.80$ \\
Context lower bound $\alpha_{\min}$ & $0.05$ \\
Context upper bound $\alpha_{\max}$ & $0.60$ \\
\bottomrule
\end{tabular}}
\end{table}

At each pruning layer, COAST first selects query-specific anchor tokens using last-token text-to-image attention. To prevent anchors from occupying the entire retained set, the number of anchors is capped by a maximum fraction of the target budget. COAST also selects low-salience contextual reference tokens using global text-to-image attention. These reference tokens are used only for contrastive scoring and do not require additional model parameters.

The entropy-guided split controls the fraction of the remaining budget assigned to complementary spatial context. Specifically, the context fraction is interpolated between $\alpha_{\min}$ and $\alpha_{\max}$ according to the normalized entropy of global text-to-image attention. This means that COAST keeps the total target budget fixed, but adapts the composition of retained tokens for each input.

\begin{figure*}[t]
    \centering
    \includegraphics[width=\linewidth]{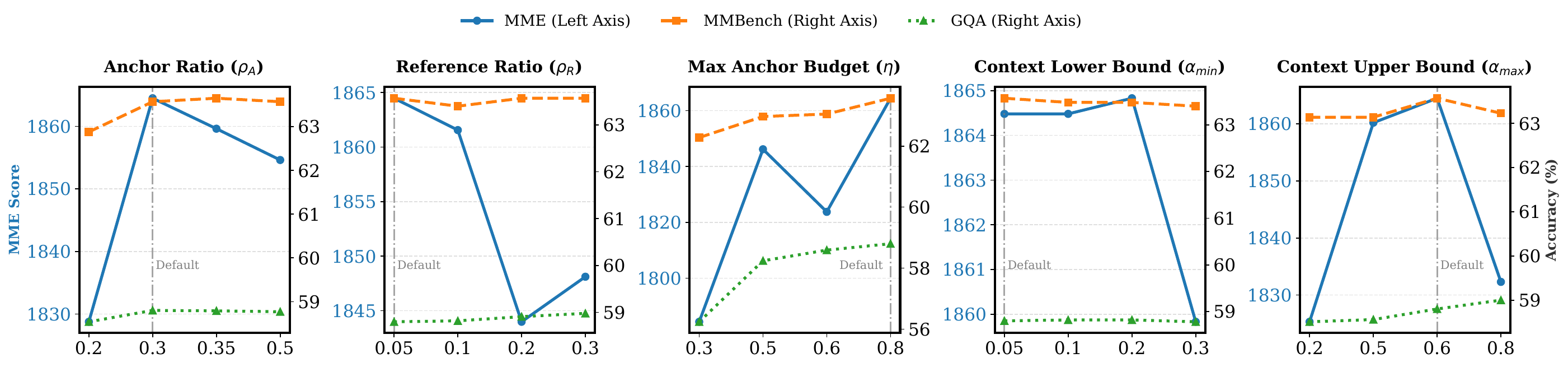} 
\caption{Hyperparameter sensitivity analysis of COAST.
We vary five routing hyperparameters while keeping the pruning budget and other settings fixed. MME is plotted on the left axis, while MMBench and GQA are plotted on the right axis. Vertical dashed lines mark the default values used in the main experiments. COAST remains stable around the default configuration, which provides a robust trade-off across the evaluated benchmarks.}
\label{fig:hyperparam_sensitivity}
\end{figure*}

\paragraph{Hyperparameter sensitivity.}
We further analyze the sensitivity of COAST to its routing hyperparameters. As shown in~\cref{fig:hyperparam_sensitivity}, we vary one hyperparameter at a time while keeping the pruning budget and all other settings fixed. Overall, COAST remains stable around the default configuration. For the anchor ratio $\rho_A$, using too few anchors degrades semantic preservation, reducing MME from $1864.48$ at the default setting to $1828.77$ when $\rho_A=0.2$. The reference ratio $\rho_R$ shows a mild trade-off: larger reference sets slightly improve GQA, while the default value achieves the best MME and competitive MMBench. The maximum anchor budget $\eta$ is more influential, where increasing it to the default value $0.8$ improves all three metrics, suggesting that sufficient anchor capacity is important under aggressive compression. The entropy bounds $\alpha_{\min}$ and $\alpha_{\max}$ are comparatively stable, with the default values providing a strong overall balance across MME, MMBench, and GQA. These results indicate that COAST is not overly sensitive to a narrow hyperparameter choice, and that the default configuration offers a robust cross-benchmark trade-off.

\subsection{Efficiency Measurement}

We report inference efficiency using both token reduction and estimated FLOPs. Token reduction is computed from the number of visual tokens retained at each decoder layer. FLOPs are estimated layer-wise using the actual sequence length after pruning. For a decoder layer with total sequence length $n$, hidden size $d$, and feed-forward intermediate size $m$, we approximate the dominant FLOPs as
\begin{equation}
    \mathrm{FLOPs}_{\mathrm{layer}}(n)
    \approx
    8nd^2 + 4n^2d + 6ndm .
\end{equation}
The first two terms correspond to attention projections and attention computation, and the last term corresponds to the feed-forward network. We sum this estimate over all decoder layers using the retained sequence length at each layer.

\subsubsection{Latency--Performance Pareto Analysis}
\label{app:pareto}

\begin{figure}[t]
\centering
\includegraphics[width=0.6\linewidth]{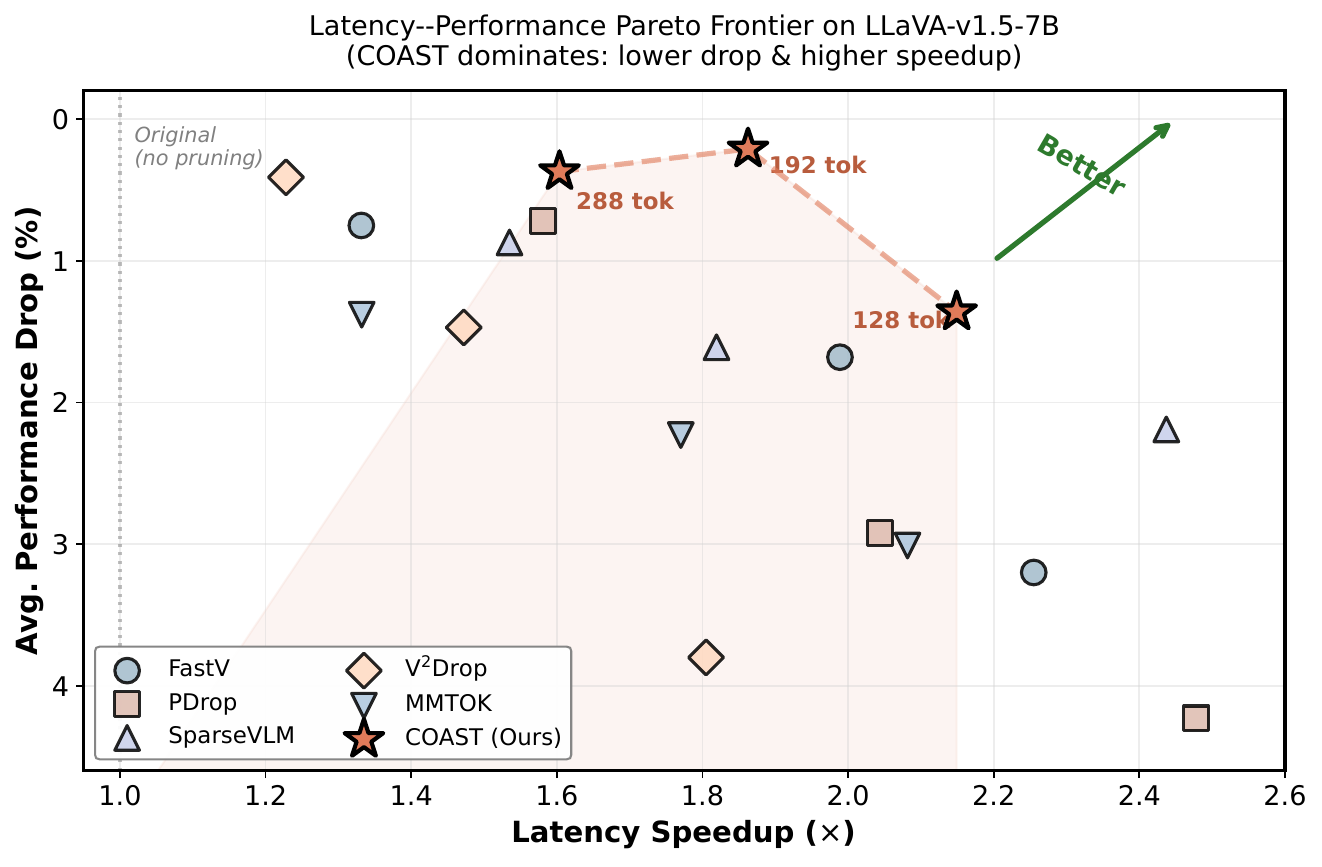}
\caption{Latency--performance Pareto frontier on LLaVA-v1.5-7B. 
We plot the latency speedup (relative to the dense baseline of $167.50$\,ms) 
against the average performance drop across seven benchmarks. 
COAST (orange stars) forms a Pareto frontier in the upper-right region 
(higher speedup, lower drop), strictly dominating all baselines: no competing 
method achieves both higher speedup and lower drop than any COAST configuration. 
Under the moderate (192-token) and aggressive (128-token) budgets, COAST 
achieves the best accuracy--efficiency trade-off, with the 192-token setting 
delivering a $1.86\times$ speedup at only $0.21\%$ performance drop.}
\label{fig:pareto}
\end{figure}

To make the accuracy--efficiency trade-off across pruning methods directly comparable, we visualize all configurations on a Pareto plot in~\Cref{fig:pareto}. Each point reports the average performance drop (across the seven benchmarks in~\Cref{tab:main_results}) versus the latency speedup relative to the dense LLaVA-v1.5-7B baseline ($167.50$\,ms). The vertical axis is inverted so that points closer to the upper-right corner are simultaneously faster and more accurate.

\paragraph{COAST forms the Pareto frontier.}
The three COAST configurations (128, 192, and 288 retained tokens) jointly trace the upper-right boundary of the achievable trade-off region. No baseline configuration achieves both higher latency speedup and lower performance drop than any COAST point: every competing method is either slower at the same accuracy or less accurate at the same speed. This Pareto dominance holds across all three compression budgets, indicating that the advantage of COAST is not tied to a particular operating point.

\paragraph{Why latency rather than FLOPs.}
Although a few baselines (e.g., MMTOK at 128 tokens, $3.53\times$ FLOPs reduction) report slightly higher theoretical FLOPs reduction than COAST, this does not translate into lower end-to-end latency: MMTOK runs at $2.08\times$ latency speedup whereas COAST achieves $2.15\times$. We attribute this gap to two factors. First, COAST's contrastive routing operates on compact anchor and reference sets, with a per-layer overhead of $O(|\mathcal{C}|(|\mathcal{A}|+|\mathcal{R}|)d)$ that is fully parallelizable and well-aligned with attention kernels. Second, COAST sorts retained tokens in their original order, preserving the contiguous layout used by subsequent decoder layers. As a result, even when COAST's FLOPs reduction is comparable to baselines, its measured latency is consistently lower, and we then view latency as the more faithful indicator of practical inference cost.

\paragraph{Behavior under aggressive compression.}
The advantage of COAST is most pronounced under aggressive compression. At 128 retained tokens (a $77.8\%$ reduction), competing methods either sacrifice substantial accuracy for speed (e.g., PDrop, $2.48\times$ speedup with $4.23\%$ drop) or sacrifice speed to preserve accuracy (e.g., SparseVLM, $2.44\times$ speedup with $2.19\%$ drop). COAST achieves $2.15\times$ speedup with only $1.36\%$ drop, simultaneously dominating both extremes. This behavior is consistent with the design intuition of two-tail retention: anchor-aligned evidence preserves query-relevant accuracy, while complementary spatial context prevents the catastrophic grounding failures (Visual Aphasia) that drive the large drops observed in baselines under tight budgets.

\subsubsection{Extended Efficiency Analysis}
\label{sec:extended_efficiency}

To comprehensively evaluate the efficiency of our method for real-world deployment, we further analyze the generation throughput, time to first token (TTFT), and peak GPU memory across different visual token budgets. As shown in Table~\ref{tab:extended_efficiency}, COAST significantly accelerates the inference process of the original LLaVA-v1.5-7B model. All efficiency metrics are evaluated on a single NVIDIA RTX 3090 GPU to ensure a fair and standardized hardware baseline.

Specifically, as the token budget decreases, the throughput improves substantially, rising from $14.07$ tokens/s in the original model to $22.06$ tokens/s under the 128-token setting (a $\sim$56.8\% speedup). Concurrently, the time to first token drops noticeably from $0.25$s to $0.15$s, ensuring a much faster initial response. Furthermore, our method consistently reduces the peak GPU memory footprint. These results further demonstrate that COAST not only preserves strong performance but also offers highly practical efficiency gains for actual deployment.

\begin{table}[h]
\centering
\small
\caption{Extended Efficiency Analysis on LLaVA-v1.5-7B. We report the generation throughput, time to first token (TTFT), and peak GPU memory footprint. COAST consistently improves all efficiency metrics across different token retention budgets.}
\label{tab:extended_efficiency}
\resizebox{0.8\textwidth}{!}{
\begin{tabular}{l c c c}
\toprule
\textbf{Budget Setting} & \textbf{Throughput (tokens/s) $\uparrow$} & \textbf{TTFT (s) $\downarrow$} & \textbf{Peak Memory (GB) $\downarrow$} \\
\midrule
Original (LLaVA-v1.5-7B) & 14.07 & 0.25 & 14.50 \\
\midrule
COAST (288 Tokens)      & 18.44 & 0.19 & 14.16 \\
COAST (192 Tokens)      & 21.08 & 0.16 & 14.07 \\
COAST (128 Tokens)      & 22.06 & 0.15 & 14.01 \\
\bottomrule
\end{tabular}
}
\end{table}

\subsection{Datasets and Evaluation Protocol}
\label{app:datasets}

We evaluate COAST on seven diverse multimodal benchmarks. To ensure a strictly fair comparison, all evaluated methods share the same backbone, prompts, decoding settings, target token budgets, and pruning schedules. This ensures that performance differences stem solely from the token selection strategy. We follow official protocols and report standard metrics for all datasets: \textbf{MME}~\cite{mme} for comprehensive perception and cognition; \textbf{MMBench-EN}~\cite{mmb} for diverse reasoning categories; \textbf{AI2D}~\cite{AI2D} and \textbf{ScienceQA-IMG}~\cite{sqa} for diagram understanding and science-domain reasoning; \textbf{GQA}~\cite{gqa} for compositional QA and spatial relations; \textbf{VizWiz}~\cite{VizWiz} for robustness on noisy, user-captured images; and \textbf{POPE}~\cite{pope} for probing hallucination-sensitive behaviors under visual compression.

\subsubsection{Ablation Protocol}
\label{app:ablation_protocol}

We conduct ablation studies to isolate the contribution of each COAST component. Unless otherwise specified, all ablation variants use the same backbone LVLM, decoding configuration, pruning schedule, and visual-token retention budget. Therefore, performance differences mainly reflect the effect of the routing or budgeting strategy rather than different compression ratios.

\paragraph{Two-tail retention.}
This ablation evaluates whether both retained candidate groups are necessary. The Top-K baseline selects visual tokens using scalar attention scores under the same retention ratio. The variant w/o $n_2$ removes the complementary spatial-context branch and keeps only anchor-aligned candidates. The variant w/o $n_1$ removes the anchor-aligned branch and keeps only complementary context candidates. COAST keeps both groups according to the entropy-guided split.

\paragraph{Entropy-driven budgeting.}
This ablation studies whether the semantic-contextual split should be fixed or input-adaptive. Fixed-ratio variants assign a constant fraction of $K_{\mathrm{rest}}$ to complementary spatial context for every input. In contrast, COAST computes the normalized entropy of global text-to-image attention and uses it to adapt the split between $n_1$ and $n_2$ for each sample.

\paragraph{Anchor--reference contrastive routing.}
This ablation evaluates the scoring function for candidate routing. The scalar variant ranks candidates using attention scores. The MaxSim variant scores candidates only by their maximum feature similarity to semantic anchors. The full COAST variant uses the anchor--reference contrastive score $\mathrm{Sim}_A-\mathrm{Sim}_R$, where $\mathrm{Sim}_A$ is the maximum similarity to the anchor set and $\mathrm{Sim}_R$ is the mean similarity to the contextual reference set.

\subsection{Quantitative Analysis of Visual Aphasia}
\label{sec:quantitative_aphasia}

To quantitatively analyze the \textit{Visual Aphasia} phenomenon discussed in the main text, we provide a detailed performance breakdown on the Adversarial split of the POPE benchmark in \cref{tab:pope_adv}. This split serves as a strict diagnostic probe, as it deliberately queries objects that frequently co-occur with the given visual context but are actually absent.

As shown in \cref{tab:pope_adv}, standard scalar pruning (FastV) suffers a catastrophic drop under extreme compression (e.g., 76.70\% at 128 tokens, an 8.43\% drop from the baseline). This quantitative degradation precisely reflects Visual Aphasia: when complementary context is discarded, the model defaults to language-prior hallucinations. In contrast, COAST maintains robust visual grounding across all budgets. Notably, under the 288-token budget, COAST achieves 84.83\%, almost fully recovering the 85.13\% accuracy of the uncompressed dense baseline.

\begin{table}[t]
\centering
\caption{Quantitative comparison on the POPE (Adversarial) benchmark. Accuracy (\%) is reported. COAST consistently outperforms FastV across all compression budgets, demonstrating its robust capability in mitigating Visual Aphasia and suppressing language-prior hallucinations.}
\label{tab:pope_adv}
\resizebox{0.75\textwidth}{!}{
\begin{tabular}{lccc}
\toprule
\multirow{2}{*}{\textbf{Method}} & \multicolumn{3}{c}{\textbf{Retained Visual Token Budget}} \\
\cmidrule(lr){2-4}
 & Retain 128 Tokens & Retain 192 Tokens &  Retain 288 Tokens \\
\midrule
LLaVA-v1.5 (Baseline) & \multicolumn{3}{c}{85.13} \\
\midrule
FastV~\cite{Fastv} & 76.70 & 80.47 & 83.40 \\
COAST (Ours) & \textbf{81.03} & \textbf{83.47} & \textbf{84.83} \\
\bottomrule
\end{tabular}
}
\end{table}

\subsection{Statistical Evidence of Attention Recovery}
\label{app:recovery_curve}

While \cref{fig:fig1_visual_aphasia} in the main text illustrates the rise-in-attention phenomenon on individual tokens from two examples, a natural concern is whether this pattern generalizes beyond hand-picked cases. To address this, we provide a statistical analysis that aggregates attention dynamics across thousands of visual tokens spanning seven benchmarks.

\paragraph{Setup.} We perform dense (no-pruning) inference on LLaVA-v1.5-7B using the respective official prompts and cache text-to-image attention at every decoder layer. For each sample, we identify the ``FastV-discarded'' set as the bottom-$(N_v - K)$ visual tokens ranked by shallow-layer ($L=2$) attention, where $K \in \{128, 192, 288\}$ corresponds to the three retention budgets used in our main experiments and $N_v = 576$ is the number of visual tokens in LLaVA-v1.5. We then measure, at each subsequent layer $l > 2$, the fraction of these initially low-attention tokens that \emph{re-enter} the top-$K$ attention ranking at layer $l$. We refer to this fraction as the \emph{Attention Rise Ratio (ARR)}. Attention scores at each layer are computed using FastV's text-to-image max-pooling protocol to ensure a fair comparison. We aggregate ARR over $N=2{,}000$ samples randomly drawn (seed 42) from the seven benchmarks: GQA, POPE, MME, MMBench-EN, AI2D, ScienceQA-IMG, and VizWiz.

\begin{figure}[t]
\centering
\includegraphics[width=0.65\linewidth]{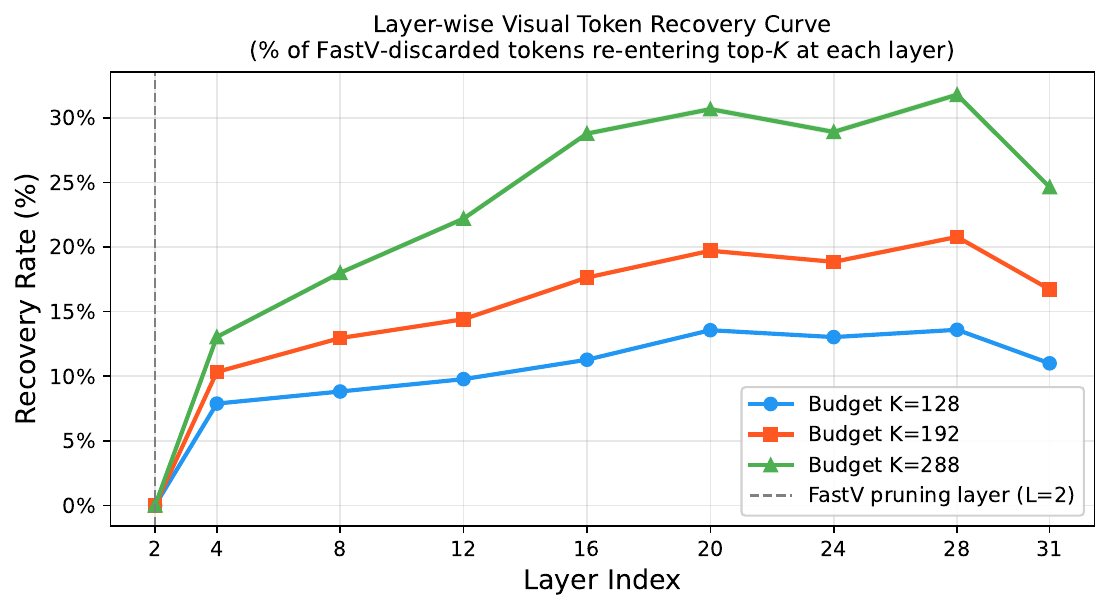}
\caption{
Layer-wise Attention Rise Ratio (ARR) on LLaVA-v1.5-7B.
For each retention budget $K \in \{128, 192, 288\}$, the curve shows the percentage of tokens (initially ranked in the bottom-$(N_v-K)$ at layer $L=2$) that re-enter the top-$K$ attention ranking at each subsequent layer. ARR rises monotonically through the network and peaks around layers 20--28, demonstrating that shallow attention substantially underestimates the visual evidence relied upon by deeper reasoning. The slight decrease at $L=31$ reflects the late-layer shift toward generation-focused attention.
}
\label{fig:recovery_curve}
\end{figure}

\paragraph{Layer-wise recovery dynamics.}
\cref{fig:recovery_curve} reports the ARR as a function of layer index for the three budgets, aggregated across all benchmarks. The recovery rate is exactly $0\%$ at the pruning layer $L=2$ by construction, but rises rapidly within the first few layers and continues to grow until layers 20--28, where it peaks at $13.5\%$, $20.3\%$, and $31.0\%$ for $K=128$, $192$, and $288$, respectively. Translated into absolute counts, this implies that FastV erroneously discards on average $\approx 60$, $\approx 78$, and $\approx 90$ semantically important tokens per sample under the three budgets. The slight decrease at $L=31$ reflects the well-documented late-layer shift toward generation-focused attention, which uniformly suppresses per-token visual attention magnitudes.

\begin{figure}[t]
\centering
\includegraphics[width=0.85\linewidth]{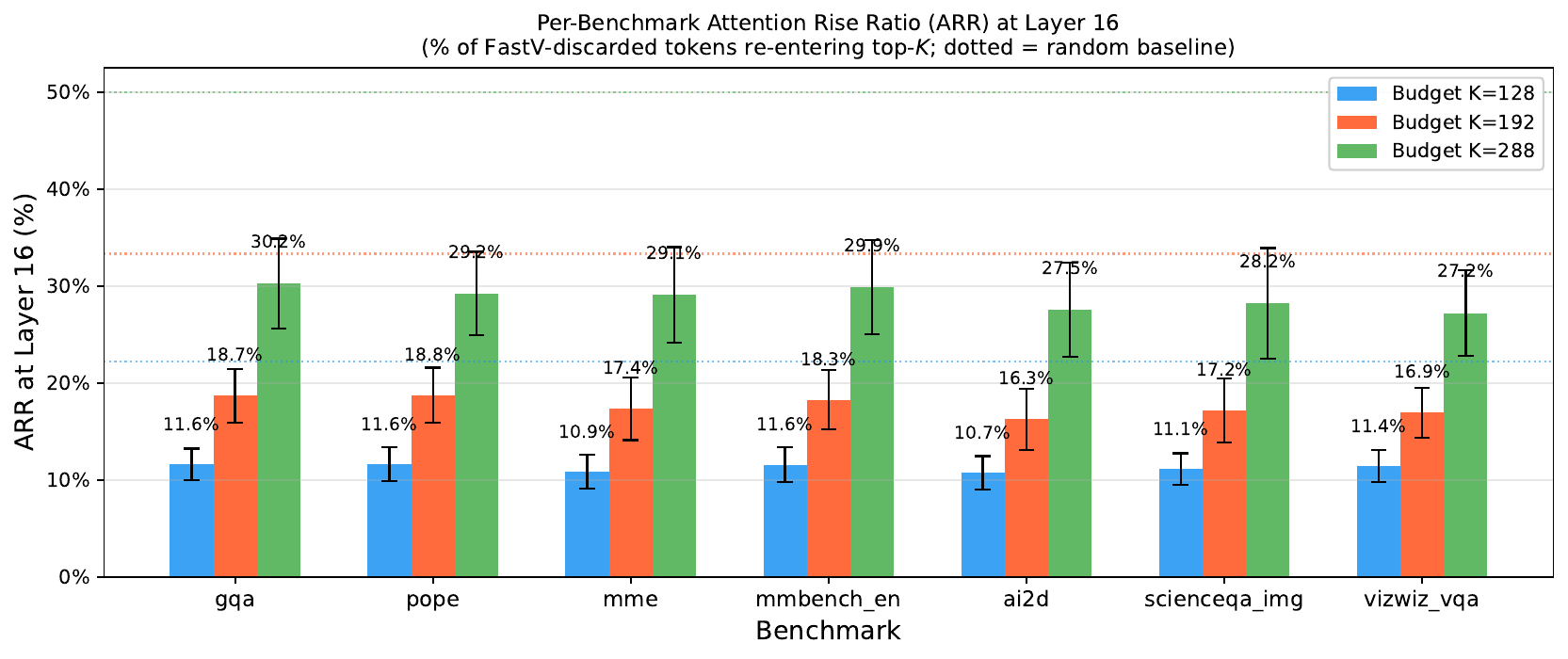}
\caption{
Per-benchmark Attention Rise Ratio (ARR) at Layer 16.
For each retention budget $K \in \{128, 192, 288\}$, bars show the percentage of FastV-discarded tokens (bottom-$(N_v{-}K)$ at layer 2) that re-enter the top-$K$ attention ranking by layer 16, averaged over each benchmark with $95\%$ bootstrap confidence intervals. Dotted horizontal lines indicate the marginal probability $K/N_v$ of any token being in the top-$K$ at random, serving as a reference for how informative FastV's shallow ranking is. ARR values are consistently in the $11$--$30\%$ range across all benchmarks, showing that the Visual Aphasia phenomenon is systematic across task types.
}
\label{fig:per_benchmark_arr}
\end{figure}

\paragraph{Cross-task generalization.}
To verify that the rise-in-attention phenomenon is consistent across task types rather than driven by any particular benchmark, \cref{fig:per_benchmark_arr} reports the layer-16 ARR separately on each of the seven benchmarks, with $95\%$ bootstrap confidence intervals. The dotted horizontal lines indicate the marginal probability $K/N_v$ of an arbitrary token being in the top-$K$ at random, providing a reference for how informative FastV's shallow ranking is.

Two observations stand out. First, the ARR remains tightly clustered across all seven benchmarks (within roughly $\pm 1\%$ for each budget), confirming that this pattern is a general property of multimodal reasoning rather than a benchmark-specific artifact. Second, ARR values are substantially below the random baseline (e.g., $\approx 11\%$ vs.\ $22.2\%$ for $K=128$), indicating that FastV's shallow ranking does carry useful signal --- yet the gap between the actual ARR and zero is precisely the population of tokens whose importance cannot be determined from shallow attention alone, and which COAST is designed to recover via two-tail retention.

\paragraph{Implication for COAST.}
These results provide the empirical foundation for COAST's two-tail retention design. Since shallow attention consistently underestimates a substantial fraction of visually important tokens (between 50 and 90 tokens per sample, depending on the budget), any pruning rule based on an early scalar score is inherently susceptible to Visual Aphasia. COAST mitigates this by (i) using anchor--reference contrastive scoring, which evaluates token importance in feature space rather than via shallow attention magnitude, and (ii) explicitly reserving a portion of the budget for complementary spatial context through entropy-driven dynamic budgeting. The improvements in \cref{tab:main_results,tab:pope_adv} align with this design rationale: COAST successfully recovers these contextually emerging tokens that scalar pruning otherwise drops.

\begin{table*}[t]
\centering
\renewcommand{\arraystretch}{0.5}
\footnotesize
\caption{Detailed generalization results across diverse MLLM architectures. The relative average performance drop compared to the Original baseline is shown in parentheses, with our method's drop highlighted in \textcolor{green!60!black}{green} for readability.}
\label{tab:appendix_generalization}
\resizebox{0.85\linewidth}{!}{
\begin{tabular}{ll *{7}{c} c} 
\toprule
\textbf{Method} & \textbf{Retain Ratio} & \textbf{MME} & \textbf{AI2D} & \textbf{GQA} & \textbf{MMB$_{en}$} & \textbf{POPE} & \textbf{SciQA} & \textbf{VizWiz} & \textbf{Avg. (\%)} \\
\midrule

\rowcolor{gray!15} \multicolumn{10}{c}{\textbf{Base Architecture: LLaVA-1.5 (13B)}} \\
\midrule
Original & 100\% & 1824.16 & 59.26 & 63.29 & 68.73 & 94.46 & 72.68 & 56.13 & 100.00 \\
\midrule
FastV & & 1796.67 & 57.84 & 60.45 & 67.27 & 95.73 & 73.97 & 57.14 & 99.20 ($\downarrow$ 0.80) \\
\rowcolor{gray!10}
Ours & \multirow{-2}{*}{$\sim$33.3\%} & 1820.21 & 58.84 & 62.19 & 68.04 & 94.91 & 73.53 & 56.60 & 99.83 (\textcolor{green!60!black}{$\downarrow$ 0.17}) \\
\midrule
FastV & & 1721.22 & 57.38 & 57.85 & 66.07 & 97.04 & 73.38 & 56.74 & 97.64 ($\downarrow$ 2.36) \\
\rowcolor{gray!10}
Ours & \multirow{-2}{*}{$\sim$22.2\%} & 1815.67 & 58.19 & 60.46 & 67.53 & 96.32 & 73.57 & 55.94 & 99.19 (\textcolor{green!60!black}{$\downarrow$ 0.81}) \\
\midrule

\rowcolor{gray!15} \multicolumn{10}{c}{\textbf{Base Architecture: LLaVA-NeXT-7B}} \\
\midrule
Original & 100\% & 1828.04 & 67.45 & 64.83 & 68.21 & 95.39 & 78.73 & 63.58 & 100.00 \\
\midrule
FastV & & 1750.27 & 65.93 & 62.48 & 65.81 & 95.70 & 73.23 & 64.43 & 97.29 ($\downarrow$ 2.71) \\
\rowcolor{gray!10}
Ours & \multirow{-2}{*}{$\sim$33.3\%} & 1855.06 & 65.74 & 63.21 & 67.44 & 95.84 & 72.68 & 63.87 & 98.37 (\textcolor{green!60!black}{$\downarrow$ 1.63}) \\
\midrule
FastV & & 1739.70 & 57.06 & 53.62 & 58.68 & 98.79 & 52.29 & 52.29 & 85.82 ($\downarrow$ 14.18) \\
\rowcolor{gray!10}
Ours & \multirow{-2}{*}{$\sim$22.2\%} & 1759.77 & 66.00 & 62.52 & 65.21 & 95.94 & 78.97 & 63.90 & 98.22 (\textcolor{green!60!black}{$\downarrow$ 1.78}) \\
\midrule

\rowcolor{gray!15} \multicolumn{10}{c}{\textbf{Base Architecture: LLaVA-OV-0.5B}} \\
\midrule
Original & 100\% & 1463.73 & 57.09 & 58.45 & 53.01 & 98.46 & 63.55 & 47.42 & 100.00 \\
\midrule
FastV & & 1438.03 & 53.72 & 51.53 & 49.74 & 98.04 & 65.54 & 46.41 & 96.42 ($\downarrow$ 3.58) \\
\rowcolor{gray!10}
Ours & \multirow{-2}{*}{$\sim$33.3\%} & 1476.26 & 54.37 & 52.86 & 50.26 & 97.64 & 66.09 & 45.63 & 97.25 (\textcolor{green!60!black}{$\downarrow$ 2.75}) \\
\midrule
FastV & & 1372.42 & 53.56 & 52.58 & 48.80 & 99.02 & 62.89 & 39.13 & 93.09 ($\downarrow$ 6.91) \\
\rowcolor{gray!10}
Ours & \multirow{-2}{*}{$\sim$22.2\%} & 1471.24 & 53.50 & 52.48 & 50.69 & 98.01 & 62.72 & 45.14 & 96.15 (\textcolor{green!60!black}{$\downarrow$ 3.85}) \\
\midrule

\rowcolor{gray!15} \multicolumn{10}{c}{\textbf{Base Architecture: Qwen2.5-VL-7B}} \\
\midrule
Original & 100\% & 2331.33 & 82.80 & 60.76 & 83.85 & 98.91 & 88.92 & 70.95 & 100.00 \\
\midrule
FastV & & 2334.72 & 78.79 & 60.46 & 84.02 & 98.82 & 88.49 & 65.22 & 98.05 ($\downarrow$ 1.95) \\
\rowcolor{gray!10}
Ours & \multirow{-2}{*}{$\sim$33.3\%} & 2334.39 & 81.90 & 60.45 & 84.02 & 96.15 & 87.90 & 68.83 & 98.83 (\textcolor{green!60!black}{$\downarrow$ 1.17}) \\
\midrule
FastV & & 2257.07 & 72.15 & 56.23 & 83.76 & 99.11 & 86.21 & 67.36 & 95.50 ($\downarrow$ 4.50) \\
\rowcolor{gray!10}
Ours & \multirow{-2}{*}{$\sim$22.2\%} & 2273.19 & 79.31 & 58.02 & 83.68 & 96.14 & 86.61 & 65.34 & 96.47 (\textcolor{green!60!black}{$\downarrow$ 3.53}) \\
\bottomrule
\end{tabular}
}
\end{table*}

\subsection{Detailed Generalization Results}
\label{appendix:appendix_generalization}

In the main text (see~\cref{fig:generalization_ablation})(a), we summarized average performance retention to demonstrate the generalizability of our proposed routing mechanism across diverse architectures. For more detailed evaluation, Table~\ref{tab:appendix_generalization} reports the complete per-benchmark results across all four evaluated LVLMs: LLaVA-1.5 (13B), LLaVA-NeXT (7B), LLaVA-OV (0.5B), and Qwen2.5-VL (7B). As shown in Table~\ref{tab:appendix_generalization}, our method consistently minimizes the performance drop relative to the uncompressed baseline across nearly all individual reasoning tasks, confirming its robustness and adaptability to various LVLM architectures.

\begin{algorithm}[t]
\footnotesize
\caption{COAST: Contrastive Adaptive Semantic Token Routing}
\label{alg:coast}
\textbf{Input:} Visual token sequence $\mathbf{X}_v \in \mathbb{R}^{N_v \times d}$, Global attention $S^{glo} \in \mathbb{R}^{N_v}$, Last-token attention $S^{last} \in \mathbb{R}^{N_v}$, Target budget $K_v$, Bounds $[\alpha_{min}, \alpha_{max}]$ \\
\textbf{Output:} Compressed and ordered visual token sequence $\widetilde{\mathbf{X}}_v$

\begin{algorithmic}[1]
\STATE \textcolor{gray}{\# 1. Entropy-Driven Dynamic Budgeting (Eq. 3-7)}
\STATE Normalize global attention: $p_j = S^{glo}_j / \sum_{m=1}^{N_v} S^{glo}_m$ for $j \in \{1, \dots, N_v\}$
\STATE Compute normalized attention entropy: $H = -\frac{1}{\log N_v} \sum_{j=1}^{N_v} p_j \log p_j$
\STATE Determine anchor budget: $K_A = \min(\rho_A N_v, \eta K_v)$
\STATE Determine remaining routing budget: $K_{rest} = K_v - K_A$
\STATE Compute context budget: $n_2 = \lfloor K_{rest} \left( \alpha_{min} + (\alpha_{max} - \alpha_{min})H \right) \rfloor$
\STATE Compute semantic evidence budget: $n_1 = K_{rest} - n_2$
\STATE
\STATE \textcolor{gray}{\# 2. Anchor and Reference Selection}
\STATE $\mathcal{A} \leftarrow \text{Top-}K_A \text{ tokens from } \mathbf{X}_v \text{ based on } S^{last}$
\STATE $\mathcal{R} \leftarrow \text{Bottom-}K_R \text{ tokens from } \mathbf{X}_v \text{ based on } S^{glo}$
\STATE $\mathcal{C} \leftarrow \mathbf{X}_v \setminus \mathcal{A}$ \hfill \textcolor{gray}{\# Remaining non-anchor candidate pool}
\STATE
\STATE \textcolor{gray}{\# 3. Anchor--Reference Contrastive Scoring (Eq. 8-10)}
\FOR{each candidate token $\mathbf{c}_i \in \mathcal{C}$}
    \STATE $\text{Sim}_A(\mathbf{c}_i) = \max_{\mathbf{a}_j \in \mathcal{A}} \frac{\mathbf{c}_i^\top \mathbf{a}_j}{\|\mathbf{c}_i\|_2 \|\mathbf{a}_j\|_2}$ \hfill \textcolor{gray}{\# MaxSim to semantic anchors}
    \STATE $\text{Sim}_R(\mathbf{c}_i) = \frac{1}{|\mathcal{R}|} \sum_{\mathbf{r}_k \in \mathcal{R}} \frac{\mathbf{c}_i^\top \mathbf{r}_k}{\|\mathbf{c}_i\|_2 \|\mathbf{r}_k\|_2}$ \hfill \textcolor{gray}{\# MeanSim to spatial references}
    \STATE $\text{Score}(\mathbf{c}_i) = \text{Sim}_A(\mathbf{c}_i) - \text{Sim}_R(\mathbf{c}_i)$ \hfill \textcolor{gray}{\# Contrastive routing score}
\ENDFOR
\STATE
\STATE \textcolor{gray}{\# 4. Two-tail Retention \& Ordered Recomposition (Eq. 11)}
\STATE $\mathcal{C}_{top} \leftarrow \text{Top-}n_1 \text{ tokens from } \mathcal{C} \text{ based on } \text{Score}(\mathbf{c}_i)$
\STATE $\mathcal{C}_{bottom} \leftarrow \text{Bottom-}n_2 \text{ tokens from } \mathcal{C} \text{ based on } \text{Score}(\mathbf{c}_i)$
\STATE Gather indices: $\mathcal{I}_{keep} \leftarrow \text{Idx}(\mathcal{A}) \cup \text{Idx}(\mathcal{C}_{top}) \cup \text{Idx}(\mathcal{C}_{bottom})$
\STATE Sort $\mathcal{I}_{keep}$ in ascending order \hfill \textcolor{gray}{\# Preserve original spatial sequence layout}
\STATE $\widetilde{\mathbf{X}}_v \leftarrow \mathbf{X}_v[\mathcal{I}_{keep}]$
\STATE \textbf{return} $\widetilde{\mathbf{X}}_v$
\end{algorithmic}    

\end{algorithm}

\subsection{Algorithm Pseudo-Code of COAST}
\label{sec:algorithm_pseudo_code}

To facilitate reproducibility, we provide the detailed algorithmic workflow of COAST in \cref{alg:coast}. The algorithm operates in a plug-and-play manner during the prefill stage at a scheduled pruning layer $l$. Notably, the cross-modal attention maps ($S^{glo}$ and $S^{last}$) are natively computed by the attention heads in the preceding layer and pooled without requiring any additional parameter projections or auxiliary modules. 
The feature similarity routing (Lines 14--18 in \cref{alg:coast}) is performed purely via matrix multiplication, enabling highly efficient parallel execution on modern GPUs. Furthermore, by maintaining the sorting of indices (Line 23), COAST ensures that the 2D spatial relationships and token position IDs remain intact for subsequent Transformer blocks.

\subsection{Observational Analysis of Layer-wise Feature Dynamics}
\label{sec:appendix_scheduling}

To better understand where token pruning can be applied with minimal disruption, we conduct an observational analysis of the evolution of visual representations across transformer layers, using LLaVA-v1.5-7B as an example.

\paragraph{Measuring Inter-layer Stability.}
We characterize the evolution of visual features by measuring their inter-layer consistency. For the $l$-th layer, we define a stability score:
\begin{equation}
S^{(l)} = \frac{1}{|\mathcal{V}^{(l)}|} \sum_{i \in \mathcal{V}^{(l)}}
\frac{\langle \mathbf{h}_i^{(l)},\, \mathbf{h}_i^{(l+1)} \rangle}
{\|\mathbf{h}_i^{(l)}\|_2 \cdot \|\mathbf{h}_i^{(l+1)}\|_2}.
\end{equation}
This metric captures how much the visual representation changes between consecutive layers. Higher values indicate that token features are relatively stable, while lower values suggest ongoing transformation.

\begin{figure}[t]
    \centering
    \includegraphics[width=\textwidth]{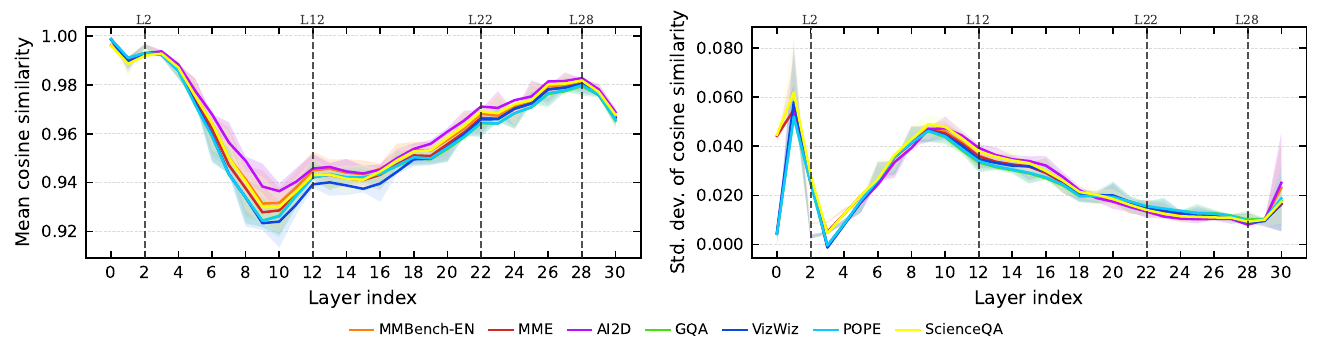}
    \caption{Layer-wise feature stability across seven benchmarks.
    Left: Mean cosine similarity between consecutive layers' visual features, $\hat{S}^{(l)}$ (Savitzky--Golay smoothed). Higher values indicate slower feature evolution.
    Right: Standard deviation of $S^{(l)}$ across visual tokens at each layer. Vertical dashed lines mark COAST's pruning layers (L2, L12, L22, L28), which correspond to the boundaries between consecutive feature-evolution phases.}
    \label{fig:layer_stability}
\end{figure}

\paragraph{Empirical Observation.}
In practice, the raw sequence $\{S^{(l)}\}$ exhibits significant local fluctuations due to architectural components such as residual connections. To reveal broader trends, we apply a 1D Savitzky--Golay filter to obtain a smoothed trajectory $\hat{S}^{(l)}$, shown as solid curves in Figure~\ref{fig:layer_stability}, while the shaded regions denote the unsmoothed standard deviation across visual tokens.
Across multiple inputs, $\hat{S}^{(l)}$ exhibits several characteristic plateaus, corresponding to stages where the rate of feature change temporarily decreases. Specifically, we observe
\textbf{(i)} a rapid reconstruction phase in shallow layers (L0--L8) where similarity drops sharply,
\textbf{(ii)} an intermediate transition phase (L9--L20) with gradual recovery, and
\textbf{(iii)} a stabilization phase in deeper layers (L20--L28) where similarity converges above $0.97$.
The standard deviation in the right panel further confirms this pattern: token differences peak during early layers and decrease as features stabilize in later layers.

\paragraph{Implications for Pruning.}
This analysis reveals that visual representations evolve unevenly across layers, with distinct phases of rapid transformation and relative stability. Guided by these transitions, COAST places pruning operations at L2, L12, L22, and L28, which correspond to the boundaries between consecutive evolution phases (Figure~\ref{fig:layer_stability}). By pruning at these boundaries, we avoid interrupting the model while features are changing rapidly. We only remove tokens after the current updates have settled.

To empirically verify this design choice, we compare three layer-selection strategies under identical retention budgets in Table~\ref{tab:layer_selection}.
\textit{Fixed-interval} places pruning at uniform layer intervals (e.g., every $\lfloor L/4 \rfloor$ layers), \textit{Random} samples pruning layers uniformly across the network, and \textit{Stability-guided} (ours) follows the phase-transition layers identified above.
The stability-guided selection consistently outperforms both baselines: it improves MME by $+102.44$ over Fixed-interval and $+80.32$ over Random, with similar gains on MMBench and GQA.
This confirms that pruning-layer placement matters, and that respecting the natural phases of feature evolution leads to a more reliable trade-off than uniform or random schedules.

\paragraph{Generalizability and Interpretability.}
Adapting COAST to new architectures (e.g., Qwen2.5-VL) simply involves running this stability analysis as a one-time, offline diagnostic step. Because it requires no model training or architectural modifications, it fully preserves the plug-and-play nature of our method during actual online inference. More importantly, this analysis serves as an interpretable tool for vision-language compression, allowing researchers to transparently discover the natural feature boundaries of any given LVLM before pruning.

\begin{table}[t]
\centering
\caption{Ablation on pruning-layer selection strategies.
All variants share the same backbone, retention budget, and number of pruning operations. \textit{Fixed-interval} places pruning operations at uniform layer intervals; \textit{Random} samples pruning layers uniformly; \textit{Stability-guided (Ours)} selects layers at the boundaries between feature-evolution phases identified in \cref{fig:layer_stability}. Best results are highlighted in \textbf{bold}.}
\label{tab:layer_selection}
\resizebox{0.6\textwidth}{!}{
\begin{tabular}{l lll}
\toprule
\textbf{Layer Selection Strategy} & \textbf{MME} & \textbf{MMB$_{en}$} & \textbf{GQA} \\
\midrule
Fixed-interval                          & 1762.04 & 60.65 & 56.57 \\
Random                            & 1784.16 & 62.71 & 57.78 \\
\midrule
\textbf{Stability-guided (Ours)}  & \textbf{1864.48} \textcolor{green!60!black}{\footnotesize{(+102.44)}} & \textbf{63.57} \textcolor{green!60!black}{\footnotesize{(+2.92)}} & \textbf{58.80} \textcolor{green!60!black}{\footnotesize{(+2.23)}} \\
\bottomrule
\end{tabular}}
\end{table}

\subsection{Limitations and Future Work}

COAST relies on native cross-modal attention as a lightweight proxy for token importance and contextual dispersion. This design makes the method training-free and easy to integrate, but its effectiveness depends on the quality of attention signals produced by the underlying LVLM. When early attention is poorly calibrated or highly sensitive to prompt formatting, the selected anchors and contextual references may be suboptimal. Future work could improve the routing signal by incorporating attention calibration, feature stability, or uncertainty estimates.

COAST also uses a predefined pruning schedule and target visual-token budget. Although entropy-guided allocation adapts the composition of the retained set, the total compression ratio remains manually specified. A natural direction is to adapt both the pruning layer and the number of retained tokens according to input difficulty, reasoning stage, or deployment constraints.

Finally, this work focuses primarily on image-based LVLM inference. Extending COAST to multi-image, video, and long-context multimodal settings may require mechanisms for temporal consistency, cross-frame token reuse, and memory-aware routing. We also consider direct diagnostics of \textit{Visual Aphasia}, such as cross-modal alignment degradation and visual evidence utilization, an important direction for better understanding pruning-induced failures.

\begin{figure}[!htbp]
\centering
\includegraphics[width=0.90\linewidth]{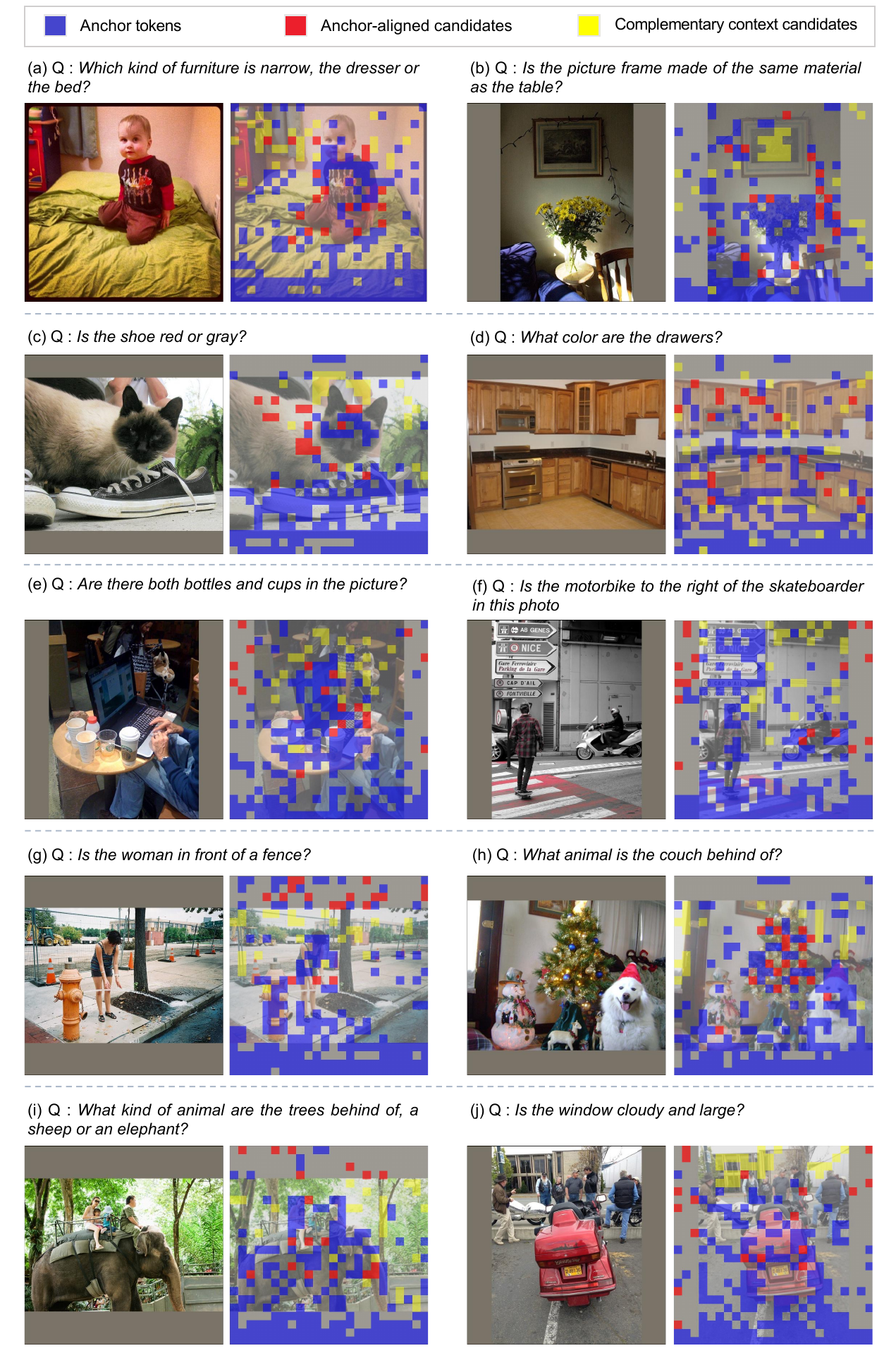}
\caption{
Visualization of COAST's Two-Tail Semantic Routing. Blue patches represent semantic anchors identified by last-token attention. Red patches represent the top-$n_1$ anchor-aligned candidates, which capture fine-grained details of the queried subjects. Yellow patches represent the bottom-$n_2$ complementary context candidates. Notice how the yellow patches naturally scatter across the background, preserving crucial spatial layouts, boundaries, and secondary visual elements that scalar pruning typically discards. Best viewed in color.
}
\label{fig:routing_visualization}
\end{figure}

\begin{figure*}[!htbp]
\centering
\includegraphics[width=0.9\linewidth]{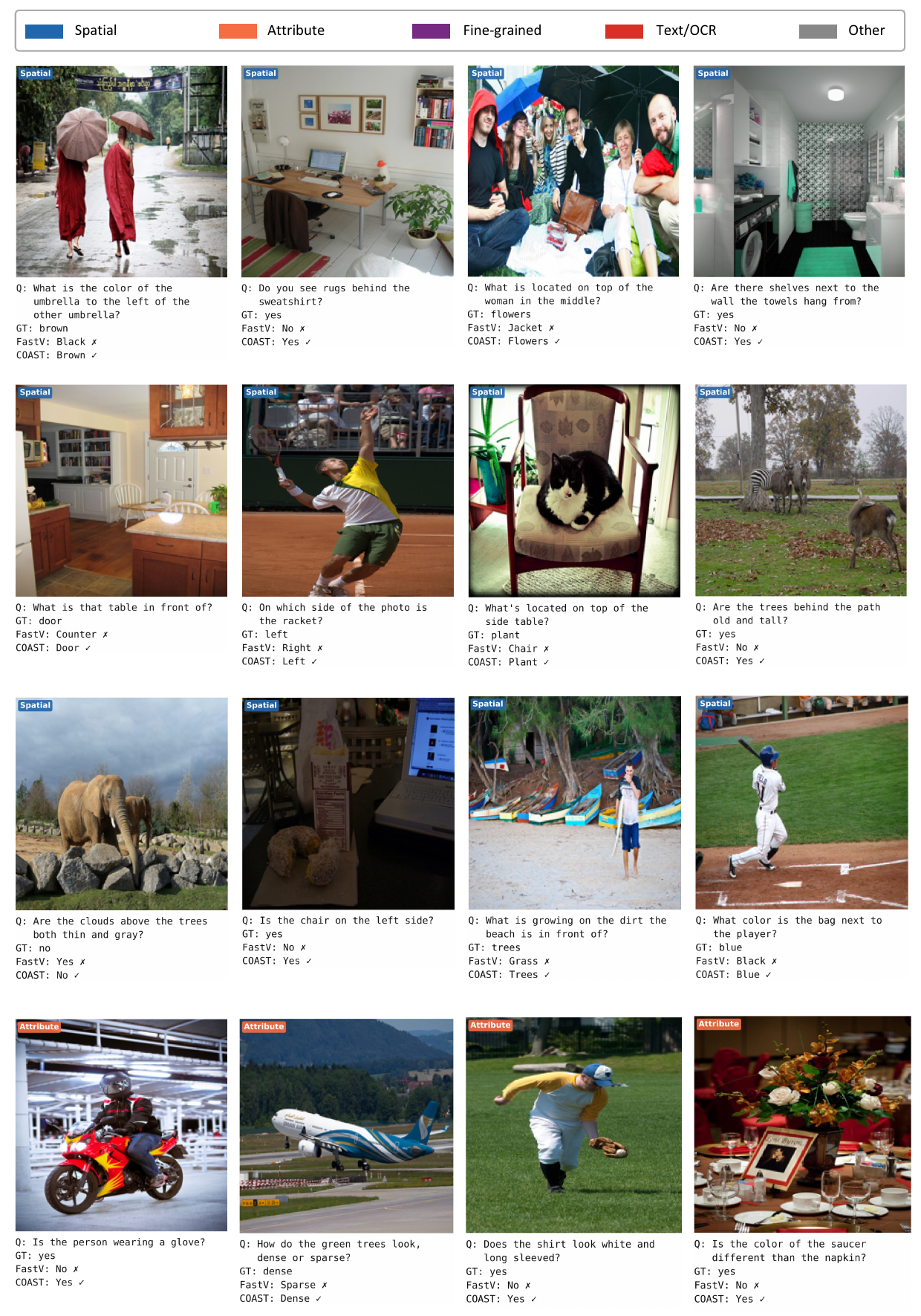}
\caption{
Extensive qualitative evidence of mitigating Visual Aphasia. 
We compare COAST against FastV on challenging queries that heavily rely on spatial reasoning and contextual grounding. FastV frequently suffers from Visual Aphasia: it prunes the vital contextual boundaries in early layers, forcing the model to guess incorrectly or hallucinate. By preserving the anchor-aligned and complementary spatial context, COAST maintains a holistic understanding of the scene, yielding correct answers.
Best viewed in color.
}
\label{fig:aphasia_case_appendix}
\end{figure*}

\begin{figure*}[!htbp]
 \ContinuedFloat
\centering
\includegraphics[width=0.9\linewidth]{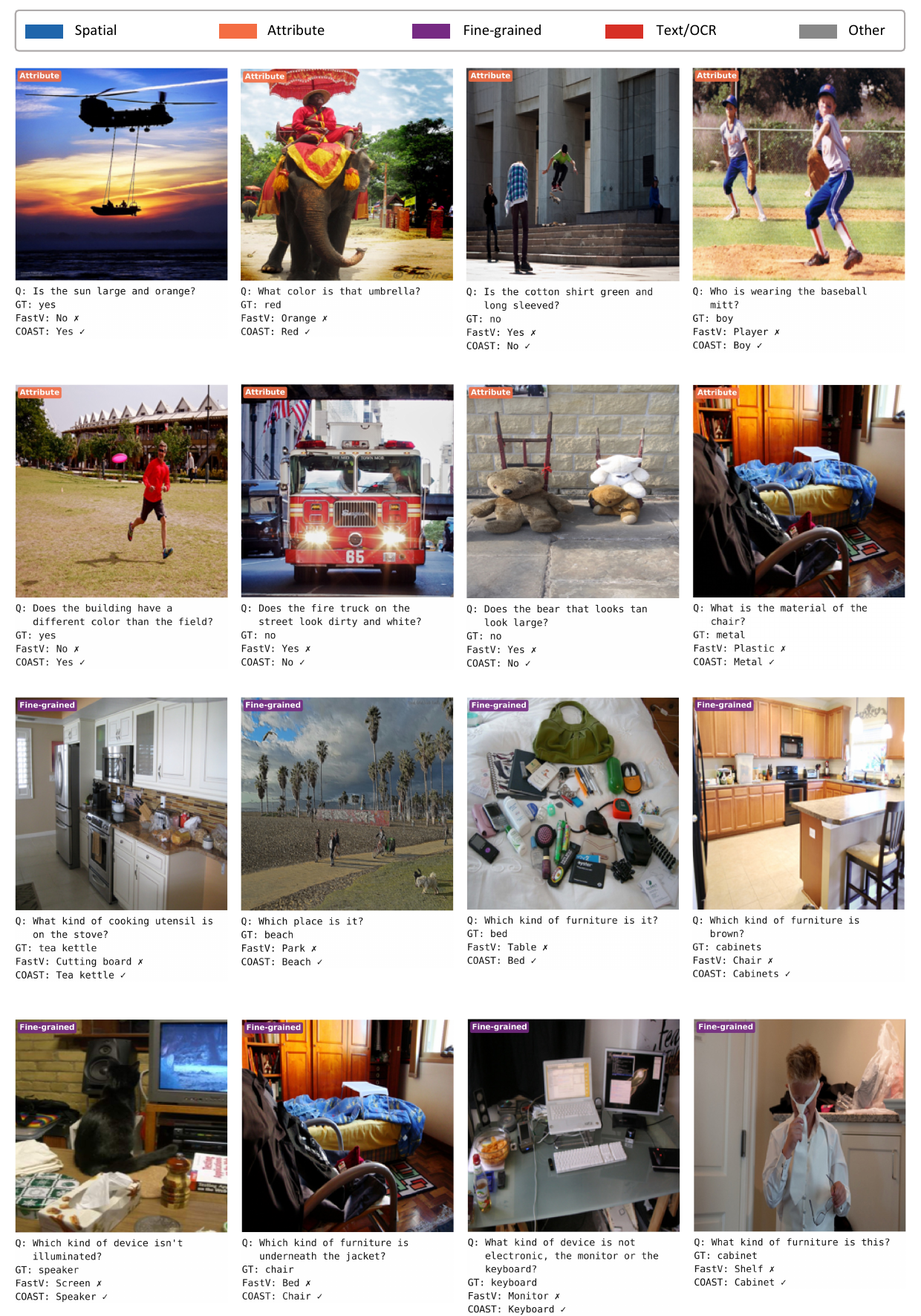}
\caption{Qualitative evidence of mitigating Visual Aphasia (Continued). Best viewed in color.
}
\end{figure*}

\begin{figure*}[!htbp]
 \ContinuedFloat
\centering
\includegraphics[width=0.9\linewidth]{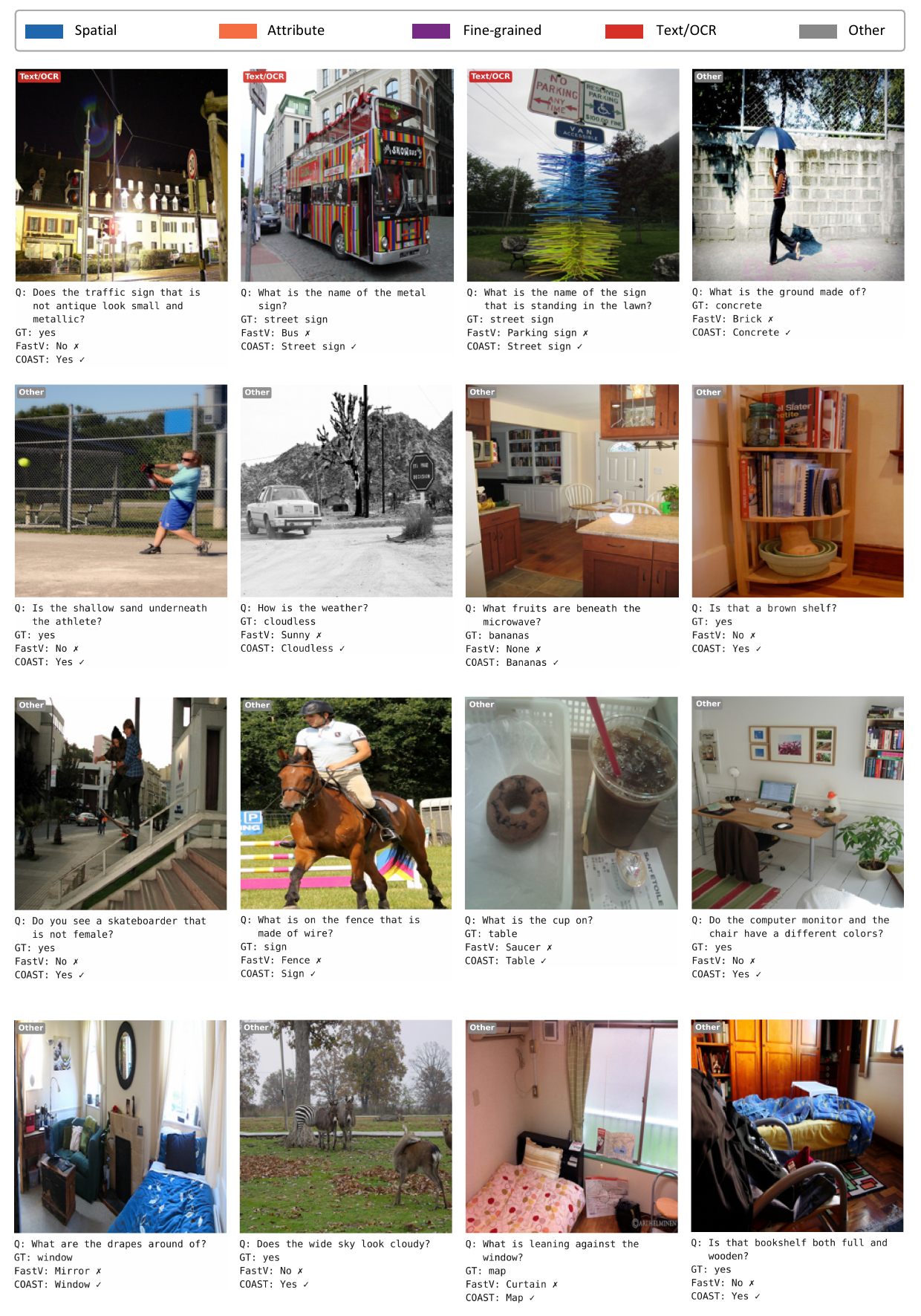}
\caption{
Qualitative evidence of mitigating Visual Aphasia (Continued).
Best viewed in color.
}
\end{figure*}

\subsection{Qualitative Analysis of Visual Aphasia}
\label{sec:appendix_visualization}

To provide an intuitive understanding of how COAST operates and why it succeeds where scalar pruning fails, we present extensive visualizations of both the token routing mechanism and the downstream question-answering outcomes.

\paragraph{How COAST routes tokens ($n_1$ vs. $n_2$).}
\cref{fig:routing_visualization} visualizes the token composition retained by COAST. Instead of preserving only highly attended regions, COAST retains both anchor-aligned evidence and complementary contextual regions. Query-specific anchors (blue) localize the primary subjects, while the top-$n_1$ anchor-aligned candidates (red) further preserve semantically relevant details around them. In addition, the bottom-$n_2$ complementary candidates (yellow) often capture broader scene context, spatial boundaries, and secondary objects that may remain useful for downstream reasoning. Together, these two groups preserve both local semantic evidence and global contextual structure during token reduction.

\paragraph{Mitigating Visual Aphasia.}
\cref{fig:aphasia_case_appendix} presents representative examples illustrating the importance of preserving complementary spatial context. In queries requiring spatial reasoning (e.g., identifying objects ``behind'', ``next to'', or ``to the left of'' another), scalar pruning methods such as FastV may remove contextual regions that are not immediately salient but remain useful for downstream reasoning. As a result, the model can produce incorrect predictions despite retaining foreground objects. 
In contrast, by preserving complementary context ($n_2$) alongside anchor-aligned evidence, COAST maintains more reliable visual grounding and achieves more accurate predictions across these challenging cases.

\end{document}